\documentclass{article}

\usepackage[nonatbib,preprint]{neurips_2021}

\usepackage[utf8]{inputenc} % allow utf-8 input
\usepackage[T1]{fontenc}    % use 8-bit T1 fonts
\usepackage{hyperref}       % hyperlinks
\usepackage{url}            % simple URL typesetting
\usepackage{booktabs}       % professional-quality tables
\usepackage{amsfonts}       % blackboard math symbols
\usepackage{nicefrac}       % compact symbols for 1/2, etc.
\usepackage{microtype}      % microtypography
\usepackage{xcolor}         % colors

\usepackage{kotex}
\usepackage{mathtools}
\usepackage{wrapfig}
\usepackage{float}
\usepackage{multirow}
\usepackage{caption}
\usepackage{tikz}
\usepackage{ctable}
\usepackage{footnote}
\newcommand*\circled[1]{\tikz[baseline=(char.base)]{\node[shape=circle,draw,inner sep=1pt] (char) {\footnotesize \textcolor{black}{#1}};}}

\title{Training Energy-Efficient Deep Spiking Neural Networks with Time-to-First-Spike Coding}

\author{%
    Seongsik Park\textsuperscript{1,2}, Sungroh Yoon\textsuperscript{1,2,3}\thanks{corresponding author} \\
    \textsuperscript{1}Department of Electrical and Computer Engineering \\
    \textsuperscript{2}Institute of New Media and Communications \\
    \textsuperscript{3}ASRI and Interdisciplinary Program in Artificial Intelligence\\
    Seoul National University, 
    Seoul 08826, South Korea \\
    \texttt{pss015@snu.ac.kr, sryoon@snu.ac.kr} \\
}

\begin{document}

\maketitle

\begin{abstract}

The tremendous energy consumption of deep neural networks (DNNs) has become a serious problem in deep learning.
%Spiking neural networks (SNNs), which mimic the operations in the human brain, have been studied as promising neural networks for energy efficiency.
Spiking neural networks (SNNs), which mimic the operations in the human brain, have been studied as prominent energy-efficient neural networks.
Due to their event-driven and spatiotemporally sparse operations, SNNs show possibilities for energy-efficient processing.
To unlock their potential, deep SNNs have adopted temporal coding such as time-to-first-spike (TTFS) coding, which represents the information between neurons by the first spike time.
With TTFS coding, each neuron generates one spike at most, which leads to a significant improvement in energy efficiency.
Several studies have successfully introduced TTFS coding in deep SNNs, but they showed restricted efficiency improvement owing to the lack of consideration for efficiency during training.
To address the aforementioned issue, this paper presents training methods for energy-efficient deep SNNs with TTFS coding.
We introduce a surrogate DNN model to train the deep SNN in a feasible time and analyze the effect of the temporal kernel on training performance and efficiency.
Based on the investigation, we propose stochastically relaxed activation and initial value-based regularization for the temporal kernel parameters.
In addition, to reduce the number of spikes even further, we present temporal kernel-aware batch normalization.
With the proposed methods, we could achieve comparable training results with significantly reduced spikes, which could lead to energy-efficient deep SNNs.

\end{abstract}

\section{Introduction} \label{sec:intro}

% deep learning 좋은 성능 but energy consumption 심각
Deep learning has shown remarkable performances in various applications, but it demands considerable energy consumption, which has become one of its biggest issues.
The energy consumption of deep neural networks (DNNs) hinders the deployment of DNN models into resource-constrained platforms, such as mobile devices, and even causes serious environmental problems~\cite{strubell2019energy}.
To improve energy efficiency, several approaches have been studied, including quantization~\cite{han2015deep,park2018quantized,hubara2017quantized,jung2019learning}, pruning~\cite{han2015learning}, and knowledge distillation~\cite{hinton2015distilling}.
These methods effectively reduce DNN model size and required amount of computations with negligible accuracy degradation, but their improvement on energy efficiency is limited by conventional computing architectures.

% SNN - intro
As an emerging computing paradigm, neuromorphic computing, which mimics the behavior of the human brain, has gained attention for its potential to improve energy efficiency further~\cite{roy2019towards}.
Neuromorphic computing is based on spiking neural networks (SNNs), which are considered the third generation of artificial neural networks (ANNs)~\cite{maass1997networks}.
SNNs consist of spiking neurons, which commonly have an integrate-and-fire feature, and the information in SNNs is represented in the form of binary spike trains, which leads to event-driven computation. 
Due to these features, deep SNNs have been regarded as prominent energy-efficient ANNs that can replace DNNs in various applications~\cite{pfeiffer2018deep,davies2021advancing}.

% SNN + training (training algorithm)
%
%하지만snn 은 학습이 어렵다는 단점
%컨버전으로 높은 정확도
% 하지만 dnn-to-snn converions의 한계(DNN에서 SNN의 특성, 효율 고려 안됨)로 인해 conversion의 효율이 제한적임 + direct training의 고효율 보고됨 (roy 논문)
Despite SNNs' potential for energy efficiency, the application of deep SNNs has been limited by the lack of scalable training algorithms.
SNNs' non-differentiable features hinder the introduction of successful DNN training algorithms, such as stochastic gradient descent (SGD) and error backpropagation, into deep SNNs.
%To overcome this limitation, various approaches have been proposed, including DNN-to-SNN conversion~\cite{diehl2015fast,rueckauer2018conversion,kim2019spiking,kim2018deep,sengupta2019going,han2020rmp,han2020deep,park2019fast,park2020t2fsnn} and surrogate training methods~\cite{mostafa2017supervised,wu2019direct,kim2020unifying,zhou2021temporal}.
To overcome this limitation, various approaches have been proposed, including surrogate training methods~\cite{mostafa2017supervised,wu2019direct,kim2020unifying,zhou2021temporal}.
%The conversion methods, which are known as indirect training, has advantages in applicability, but is limited in efficiency because they cannot consider features of SNNs during training~\cite{lee2020enabling}.
%On the contrary, direct training with the surrogate model using approximate differentiable spiking function enables efficiency-aware training of deep SNNs.
These approach using approximate differentiable spiking function enables efficiency-aware training of deep SNNs.
% SNN + temporal information
In another approach to fully utilizing energy-efficient potentials, deep SNNs have adopted temporal coding, such as time-to-first-spike (TTFS)~\cite{zhang2019tdsnn,park2020t2fsnn,zhou2021temporal}, which represents the information with spike time and has shown superior efficiency with fewer spikes.
%Among them, TTFS coding, which represents the information with spike time, has shown superior efficiency with the fewest spikes.
However, their methods in deep SNNs have been limited due to the lack of consideration for efficiency during training.
Thus, for energy-efficient deep SNNs, it is required to train deep SNNs with TTFS coding considering efficiency.

% in this work, SNN + temporal + training
% important integration
In this paper, to achieve energy-efficient deep SNNs, we introduce a scalable training algorithm, based on supervised learning with SGD, for deep SNNs with TTFS coding.
For training deep SNNs in a feasible time, we present a surrogate model, which operates in the spike time domain and has a clipped activation function.
The surrogate model has pruned and quantized activation because of restrict time window and integer spike time, respectively.
We investigated the impact of temporal kernel on training performance and efficiency in the surrogate model.
Based on the analysis, we found that gradients were vanished depending on parameters in the temporal kernel, which is used for TTFS coding.
To resolve the gradient vanishing, we propose stochastically relaxed activation.
In addition, we introduce temporal kernel regularization for stable training and for reducing the number of spikes in deep SNNs.
To further reduce the number of spikes, we propose temporal kernel-aware batch normalization.
We evaluated the proposed methods with deep SNNs, which are based on VGG-16~\cite{simonyan2014very}, on the CIFAR datasets~\cite{Krizhevsky09learningmultiple}.
Our contributions can be summarized as follows:
\begin{itemize}
\item \textbf{Computationally Efficient surrogate DNN model:} We propose a surrogate DNN model for training a deep SNN with TTFS coding at a feasible training cost.
\item \textbf{Robust training methods:} We introduce stochastically relaxed activation and temporal kernel regularization to improve training performance of the proposed surrogate model.
\item \textbf{Significant reduction in the number of spikes:} We propose temporal kernel-aware batch normalization to reduce the number of spikes in deep SNNs with TTFS coding.
\end{itemize}

\section{Related Work}

\subsection{Spiking Neural Networks} \label{sec:snn}

% introduction to SNN
SNNs consist of spiking neurons, which integrate incoming information into their internal states and generate binary spikes only when the state exceeds a certain threshold.
This integrate-and-fire feature of spiking neurons induces spatiotemporally sparse activations in SNNs.
Each spiking neuron communicates subsequent neurons with a set of binary spikes; this is called spike train.
This feature allow SNNs to be multiplication-less neural networks and exploit event-driven computing.
Due to the above characteristics, SNNs have been considered as promising for energy efficiency and as the third-generation ANNs~\cite{maass1997networks}.

% IF model - general formulation of SNN
There have been various types of spiking neurons~\cite{izhikevich2004model}. 
Among them, simplified spiking neuron models, such as integrate-and-fire (IF) neuron~\cite{abbott1999lapicque}, have been widely used for their low computational complexity.
Accordingly, in this work, we adopted the IF neuron for deep SNNs.
The IF neuron's integration process into a membrane potential $u$, which denotes the internal state, can be stated as
\begin{equation}
\label{eq:vmem_dyn}
    \frac{du_{j}^{l}(t)}{dt} = \sum\nolimits_{i}{w_{ij}^{l} \sum\nolimits_{f}{\kappa^{l}(t-t_{i,f}^{l\textrm{-}1})} + \eta_{j}^{l}(t)} + b_{j}^{l} \textrm{,}
\end{equation}
where $w$ is a synaptic weight; $\kappa$ is a synaptic kernel that is induced by binary input spikes; $\eta$ is a reset function; $b$ is a bias; $l$, $i$, and $j$ indicate the indices of layer, pre-, and post-synaptic neurons, respectively.
$t_{i,f}^{l}$ is the $f$th spike time satisfying the firing condition as
\begin{equation}
\label{eq:firing_time}
    t_{i,f}^{l}: u_{i}^{l}(t_{i,f}^{l}) \geq \theta_{i}^{l}(t_{i,f}^{l}) \textrm{,}
\end{equation}
where $\theta$ is the threshold.
Since the information between neurons should be represented in the sequence of binary and discrete spikes as in Eqs.~\ref{eq:vmem_dyn} and \ref{eq:firing_time}, the representation method significantly affects both the performance and efficiency of SNNs.

% neural coding
Neural coding defines the way of representing the information in the form of spike trains including encoding and decoding function~\cite{brette2015philosophy}.
%There have been various types of neural coding, such as rate~[refs], phase~[refs], burst~[refs], temporal-switching-coding (TSC)~\cite{han2020deep} and time-to-first-spike (TTFS) coding~[refs], as depicted in Fig.~\ref{fig:TTFS}.
There have been various types of neural coding, such as rate~\cite{sengupta2019going,han2020rmp,kim2019spiking}, phase~\cite{kim2018deep}, burst~\cite{park2019fast}, temporal-switching-coding (TSC)~\cite{han2020deep}, and TTFS coding~\cite{mostafa2017supervised,zhang2019tdsnn,park2020t2fsnn,zhou2021temporal}.
%It is mainly divided into two classes: rate and temporal coding.
%The rate coding utilizes spike firing rate and have been widely used in many studies for its low implementation overhead~[refs - rate coding].
%However, it has shown restricted performance and efficiency in deep SNNs due to its insufficient exploitation of temporal information in spike trains~[refs - limitations in rate coding].
%due to insufficient use of the temporal information 
%due to the lack of exploiting 
To maximize the efficiency by fully utilizing the temporal information in spike train, TTFS coding, which is known as latency coding, was introduced in SNNs~\cite{mostafa2017supervised}.
%As shown in Fig.~\ref{fig:TTFS}, the TTFS coding utilizes the first spike time and, thus, can significantly reduce the number of spikes, which results in considerable improvement in SNNs' efficiency.
The TTFS coding utilizes the first spike time and, thus, can significantly reduce the number of spikes, which results in considerable improvement in SNNs' efficiency.
Especially, reduction in spikes with TTFS coding can resolve the spike congestion problem, which is the most urgent issue for deep SNNs on neuromorphic architecture~\cite{davies2021advancing}.

%\textcolor{red}{TTFS, TSC / one spike, two spike (ISI) 차이 명시}

% TTFS + deep SNNs (implementation, noise etc)
% SNN model (neuron and synapse) for TTFS coding
% efficiency and importance of TTFS coding (importance of exploiting temporal information in SNNs, especially deep SNNs)
    % features and feasibility of temporal coding
    % extremely low spikes -> energy-efficient
    % vulnerable to noise -> some previous works to address this issue % cite DAC-21
% formulation
Recently, several studies successfully adopted TTFS coding to deep SNNs~\cite{zhang2019tdsnn,park2020t2fsnn,zhou2021temporal}.
%According to [refs - TDSNN, T2FSNN], to implement TTFS coding in deep SNNs, there should be dependencies between integration and fire phase of subsequent layers, as depicted in Fig.~\ref{fig:TTFS}-(b).
According to \cite{zhang2019tdsnn,park2020t2fsnn}, to implement TTFS coding in deep SNNs, there should be dependencies between integration and fire phases of subsequent layers.
The integration phase, which corresponds to the decoding function, can be implemented with an exponentially decaying synaptic kernel $\kappa$, which is described as
\begin{equation}
\label{eq:kernel}
    \kappa^{l}(t) = H(t-t_{\textrm{r}}^{l}) H(t_{\textrm{r}}^{l\textrm{+}1}+1-t) \exp(-({t-t_{\textrm{r}}^{l}-t_{\textrm{d}}^{l}})/{\tau^{l}}) \textrm{,}
\end{equation}
where $H$ is the Heaviside step function; and $\tau^{l}$, $t_{\textrm{d}}^{l}$, and $t_{\textrm{r}}^{l}$ are time constant, time delay, and reference time of $l$th layer, respectively.
The fire phase contains the encoding function with a time-variant threshold, which is given by $\theta^{l}(t) = \theta_{0}^{l} \kappa^{l+1}(t)$, where $\theta_{0}^{l}$ is the initial threshold of layer $l$, and there is a sufficiently long refractory period by the reset function $\eta(t)$ after the first spike generates.
%The two Heaviside step functions in Eq.~\ref{eq:kernel} define the time window $T$, which is given time steps to represent spike time in each layer, as shown in Fig.~\ref{fig:TTFS}-(b).
The two Heaviside step functions in Eq.~\ref{eq:kernel} define the time window $T$, which is given time steps to represent spike time in each layer.
%More detailed explanation is presented in Appendix~[ref].
As depicted in Eqs.~\ref{eq:vmem_dyn} and \ref{eq:kernel}, the temporal kernel determines the precision and representation range of information and, thus, has a significant impact on the information transmission between neurons.
%Therefore, it is required to optimize the kernel parameters in Eq.~\ref{eq:kernel} for accurate and efficient SNNs.

%
%\begin{equation}
%\label{eq:vth}
    %\theta^{l}(t) = \theta_{0}^{l} \kappa^{l+1}(t) \textrm{,}
%\end{equation}

% 맺음말 - prospect and current obstacles of TTFS coding

%%%%%%%%
% tmp 
%%%%%%%%

% SNN formulation and notations

%\cite{mostafa2017supervised}
%\cite{park2020t2fsnn}

\subsection{Training Methods of deep SNNs}
%\subsection{Training Methods of SNNs with TTFS coding}

% intro - training methods for SNN, importance
SNNs have non-differentiable and event-driven computing features, which make it challenging to adopt successful DNN training algorithms.
Instead of them, biologically plausible algorithms have been used to train SNNs, such as spike-timing-dependent plasticity (STDP)~\cite{diehl2015unsupervised}.
These unsupervised approaches can train SNNs, but they are limited in shallow networks and training performance.

% conversion
To overcome limitations of unsupervised learning in deep SNNs, DNN-to-SNN conversion methods were proposed~\cite{diehl2015fast,sengupta2019going,han2020rmp}.
These indirect training approaches exploit both advantages in the training performance of DNNs and energy-efficient inference of SNNs.
%In these methods, inference is conducted in the deep SNNs, which are converted from pre-trained DNNs.
With the conversion, deep SNNs achieved comparable results to DNNs in various applications, including image classification~\cite{sengupta2019going} and multi-object detection~\cite{kim2019spiking}.
%Furthermore, to improve the inference efficiency, various neural coding schemes were successfully applied to the converted SNNs [refs - neural coding, conversion].
However, conversion approaches are limited in efficiency~\cite{lee2020enabling}, which is defined with latency and the number of spikes, even if the converted SNNs adopted several temporal coding schemes, such as TTFS coding~\cite{zhang2019tdsnn,park2020t2fsnn}.

%Especially, in [refs-TTFS,conversion], deep SNNs with TTFS coding could improve the energy efficiency significantly because of considerably reduced the number of spikes.
%몇몇 연구에서는 conversion 후에 효율성을 높이는 방법을 제시하였으나, 
%However, their efficiencies were restricted by the lack of consideration for deep SNNs during training DNNs [ref - K. roy].

%
%\textcolor{red}{(add direct training methods details, surrogate gradient and model)}
Direct training methods were proposed to address the aforementioned issues in biologically plausible training algorithms and DNN-to-SNN conversion.
These approaches, which are based on supervised learning, use surrogate gradients~\cite{wu2018spatio,zheng2020going,wu2021training} or surrogate models~\cite{thiele2019spikegrad} to overcome the non-differentiability and bring DNNs' training algorithms to deep SNNs.
Surrogate gradient approaches successfully train deep SNNs with SGD, but they demand tremendous computing overhead according to the spike activations, which prevents efficient application to deep SNNs.
%By approximating the non-differentiable spiking function with surrogate gradient, they successfully trained deep SNNs with SGD and BP as in conventional DNNs [refs].
%However, these methods demanded tremendous computing overhead according to the spike activations, which hinders their efficient application in deep SNNs.
To train deep SNNs efficiently, surrogate models have been studied~\cite{thiele2019spikegrad}.
These methods have shown successful training results on deep SNNs.
Thus, in this paper, we adopted the surrogate model approach.
%
% direct training + neural coding (temporal coding)
% shallow or did not consider the efficiency, which is an important
From the perspective of neural coding, most studies have focused on rate coding, and only a few have trained SNNs with TTFS coding~\cite{mostafa2017supervised,zhang2019tdsnn,park2020t2fsnn,zhou2021temporal}.
%TTFS coding is considered as the most promising neural coding for deep SNNs due to its efficiency and has been adopted various studies [refs].
Despite the efforts to improve the efficiency of TTFS coding in deep SNNs~\cite{park2020t2fsnn,park2021noise}, their improvements have been restricted by the conversion-based training algorithms.
In several studies, SNNs were directly trained, but their methods were not validated as applicable to deep SNNs~\cite{mostafa2017supervised,goltz2019fast,comcsa2021temporal}.
A recent study suggested direct training methods for deep SNNs with TTFS coding~\cite{zhou2021temporal}.
However, few previous studies considered the efficiency during training process.

\section{Surrogate DNN Model for TTFS coding}

\begin{figure}[t]
    \centering
    \includegraphics[width=0.9\linewidth]{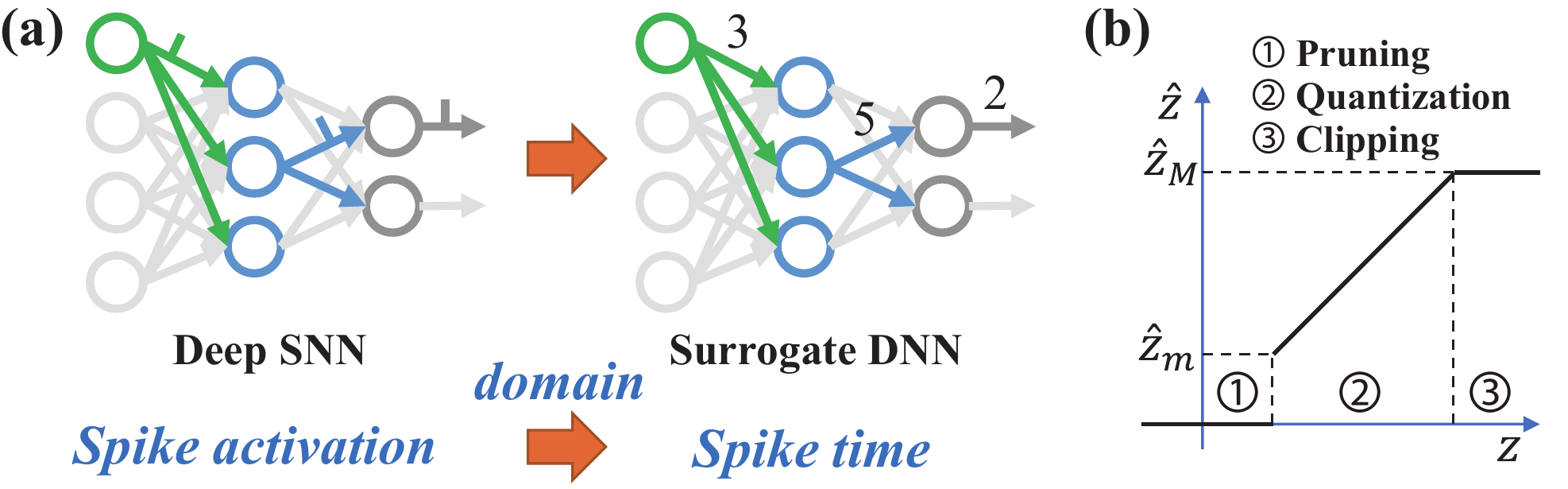}
	\vspace{-0.5em}
	\caption{Proposed surrogate DNN model for deep SNNs with TTFS coding. (a) domain transformation from spike activation domain of deep SNNs to spike time domain of surrogate DNNs. (b) activation function in the surrogate model}
	\label{fig:surrogate_dnn}
	\vspace{-1.0em}
\end{figure}

% to do
% direct training w/ surrogate gradient 문제
% need for surrogate model
SNNs have a feature of temporal processing with spiking neurons, which requires computations depending on time steps and is executed iteratively as recurrent neural networks (RNNs).
Furthermore, in TTFS coding, the inference latency is proportional to the depth of neural networks and is usually more than several hundreds time steps in deep SNNs~\cite{zhang2019tdsnn,park2020t2fsnn}.
For these reasons, the direct training with surrogate gradients of SNNs demands more than $LT$ time computations compared to DNNs, where $L$ is the number of layers and $T$ is time window of each layer.
Thus, considering the computational complexity, it is not feasible to apply the surrogate gradient approach to training deep SNNs with TTFS coding.

% explain details of proposed surrogate model
% encoding / decoding function
To address the time complexity issue, we introduce a surrogate DNN model that contains encoding and decoding functions of TTFS coding, as shown in Fig.~\ref{fig:surrogate_dnn}.
In deep SNNs with TTFS coding, the activation is represented by the first spike time.
Hence, we exploit this fact to transform the activation domain from the spike activation domain of deep SNNs to the spike time domain of the surrogate model.
%Through this transformation, we can utilize DNNs' well-defined algorithms to train deep SNNs with TTFS coding.
Through this transformation, we can utilize DNNs' well-defined algorithms to train deep SNNs.
According to Eqs.~\ref{eq:firing_time} and \ref{eq:kernel}, the first spike time in deep SNNs can be formulated as
\begin{equation}
\label{eq:encoding}
    t_{\textrm{1st}}^{l} = \lceil -\tau^{l} \ln( u_{i}^{l}(t_{\textrm{r}}^{l}-1)/\theta_{0}) + t_\textrm{d}^{l} \rceil \textrm{,}
\end{equation}
where $u_{i}^{l}(t_{\textrm{r}}^{l}-1)$ is the integrated membrane potential of the neuron.
When we map the membrane potential $u$ in SNNs to pre-activation value $z$ in the surrogate DNNs, the encoding $f$ and decoding $g$ functions are stated as
\begin{equation}
\label{eq:s_dnn_enc_dec}
    f(z^{l}) = t^{l} = R\{Q[-\tau^{l}\ln(R(z^{l})+\epsilon)+t_{\textrm{d}}]\}+t_{\textrm{r}}^{l}
    \quad\textrm{and}\quad
    g(t^{l}) = \hat{z}^{l} = \kappa^{l}(t^{l}) \textrm{,}
\end{equation}
where $R$ and $Q$ are ReLU and quantization function, respectively; $\epsilon$ is a small constant for numerical stability; and $\kappa$ is the temporal kernel, which is depicted in Eq.~\ref{eq:kernel}.
The spike time activation $t^{l}$ is transmitted and added to subsequent neurons through the decoding function.
The weighted sum of the decoded values $\hat{z}$ forms the next-layer's pre-activation as $z_{j}^{l\textrm{+}1}=\sum_{i}{w_{ij} \hat{z}_{i}^{l}}$.

% temporal kernel - pruning, quantization, clipping
%\textcolor{red}{pruning clipping - temporal kernel representation range}
In TTFS coding, earlier spikes convey more crucial information than later spikes, and spikes that occur after the time window $T$ in each layer do not affect the next layer.
We implemented these features of TTFS in the surrogate model by the encoding and decoding procedures with the spike time and the exponentially decaying temporal kernel (Eq.~\ref{eq:kernel}).
Depending on the encoding value, which is a pre-activation $z$, the decoded value $\hat{z}$ can be divided into three types, as described in Fig.~\ref{fig:surrogate_dnn}-(b): pruned, quantized, and clipped activations. 
Smaller values than the minimum representation $\hat{z}_{m}$, which correspond to the later spike times than the time window, are eliminated by the pruning.
On the other hand, larger activation values than the maximum representation value of the kernel $\hat{z}_{\textrm{M}}$ are encoded into the smallest spike time and clipped to the maximum value.
The pre-activations between $\hat{z}_{\textrm{m}}$ and $\hat{z}_{\textrm{M}}$ are quantized by the encoding function.

% end-to-end trainable, gradients
% trainable model - the surrogate model consists of differentiable functions, gradients of temporal kernel parameters
The pre-activation $z$, which corresponds the membrane potential in SNNs, is encoded in the spike time $t$, which is a positive and integer value, as an activation in the surrogate model.
With the encoding function $f$, the non-differentiable spiking function (Eq.~\ref{eq:firing_time}) is replaced by the quantization function $Q$, which can be used in end-to-end training with a straight-through estimator~\cite{hubara2017quantized}.
In addition, the pruned and clipped activations by the decoded function are also trained with the quantized values as in \cite{jung2019learning}.
The gradients of classification loss (cross-entropy loss) with respect to the temporal kernel parameters are obtained as follows:
\begin{equation}
\label{eq:grad_tk}
    \frac{\partial{L_{\textrm{CE}}}}{\partial{\tau}^{l}}
    = \frac{\partial{L_{\textrm{CE}}}}{\partial{z}^{l\textrm{+}1}} \frac{\partial{z}^{l\textrm{+}1}}{\partial{\tau}^{l}}
    = \frac{\partial{L_{\textrm{CE}}}}{\partial{z}^{l\textrm{+}1}} [w^{l}\hat{z}^{l}(\frac{t^{l}-t_{\textrm{d}}^{l}}{(\tau^{l})^2})]
    \quad\textrm{and}\quad
    \frac{\partial{L_{\textrm{CE}}}}{\partial{t_{\textrm{d}}^{l}}}
    = \frac{\partial{L_{\textrm{CE}}}}{\partial{z}^{l\textrm{+}1}} \frac{\partial{z}^{l\textrm{+}1}}{\partial{t_{\textrm{d}}^{l}}}
    = \frac{\partial{L_{\textrm{CE}}}}{\partial{z}^{l\textrm{+}1}} [w^{l}\hat{z}^{l}(\frac{t_{\textrm{d}}^{l}}{\tau^{l}})]
    \textrm{.}
\end{equation}
% 맺음
We updated the kernel parameters with the gradients as weight parameters.

\section{Improving Training Performance of the Surrogate Model} \label{sec:training_methods}

\begin{figure}[t]
    \centering
    \includegraphics[width=1.0\linewidth]{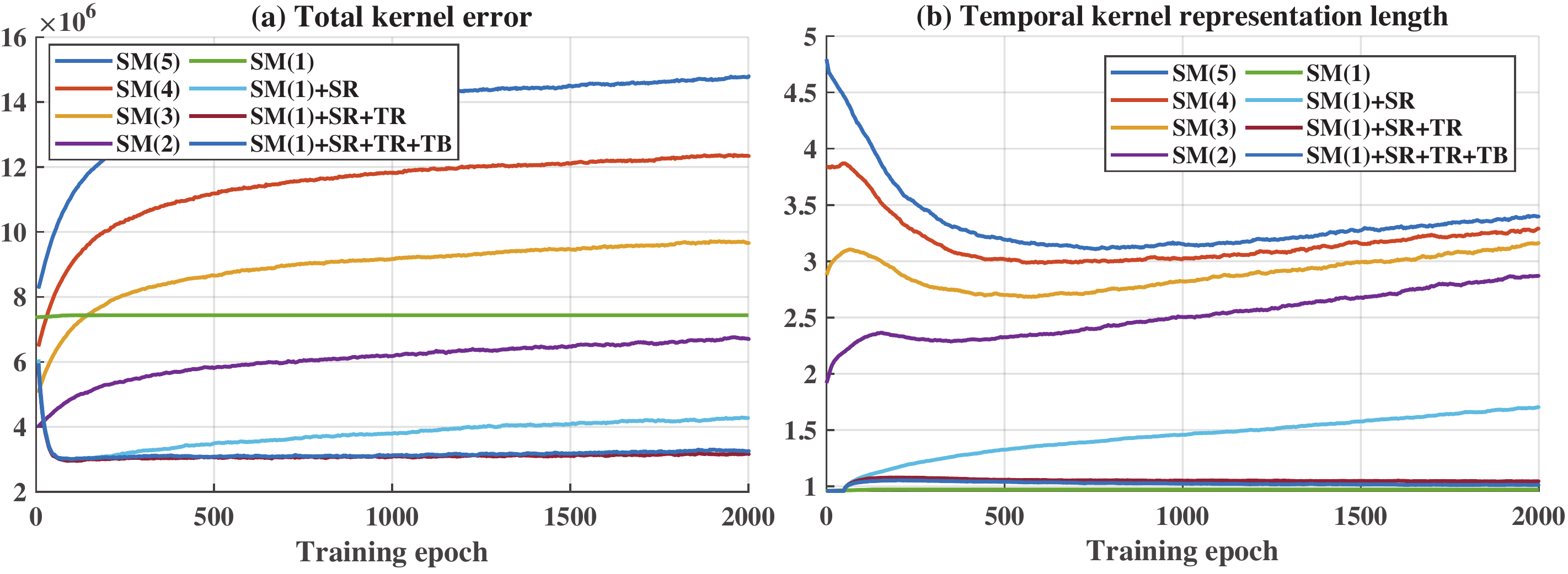}
	\vspace{-1.2em}
	\caption{(a) Error in the temporal kernel $\kappa$ and (b) representation range of the temporal kernel depending on the initial maximum kernel value $\hat{z}_{\textrm{M,0}}$, which is denoted in parenthesis, during training deep SNNs (VGG-16, CIFAR-10). SM: Surrogate Model, SR: Stochastic Relaxation, TR: Temporal kernel regularization, TB: Temporal kernel-aware batch normalization}
	\label{fig:tk_error_range}
	%\vspace{-2.0em}
\end{figure}

\begin{figure}[t]
    \centering
    \includegraphics[width=1.0\linewidth]{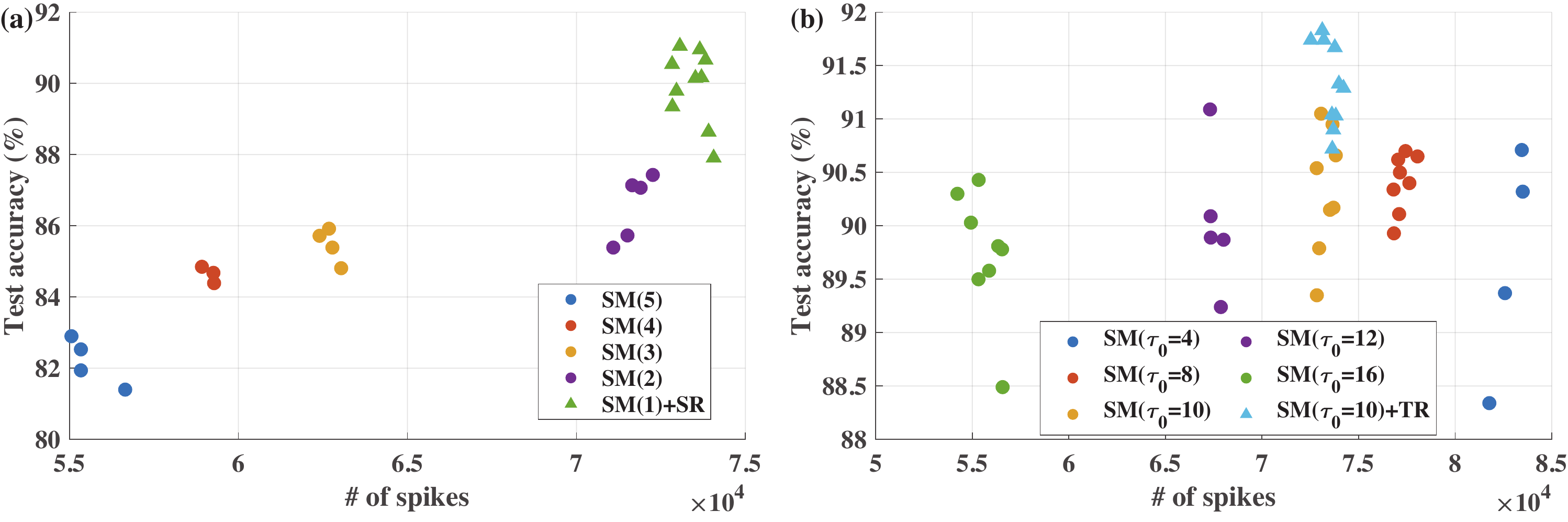}
	\vspace{-1.2em}
	\caption{Training results (test accuracy and the number of spikes) of deep SNNs (VGG-16) on CIFAR-10. 
	(a) Results depending on the initial maximum kernel values $\hat{z}_{\textrm{M,0}}$ with $\tau_{0}$=10. We omit the results of $\hat{z}_{\textrm{M,0}}$=1 due to its poor training results. The values in the parenthesis indicates $\hat{z}_{\textrm{M,0}}$. (b) Results according to various initialization of $\tau_{0}$ with $\hat{z}_{\textrm{M,0}}$=1 and stochastic relaxation.}
	\label{fig:spike_acc_sr_reg}
	\vspace{-1.0em}
\end{figure}

% problems in the surrogate model
% tradeoff in temporal kernel - representation range, precision
Due to the temporal kernel, the decoded information is represented by a certain number of quantization levels, which is defined by the time window $T$, and has a limited range, which is stated as $[\hat{z}_{\textrm{m}},\hat{z}_{\textrm{M}}]=[\exp(-T/\tau),1]\exp(t_{\textrm{d}}/\tau)$.
The length of the representation range is determined by the maximum value of the kernel ($\hat{z}_{\textrm{M}}=\exp(t_{\textrm{d}}/\tau)$); thus, the averaged quantization error, which corresponds to the representation length divided by the number of quantization level, is proportional to the maximum value.
%In addition, the minimum representable value of the kernel is defined by the maximum representation.
Hence, there is a trade-off between representation range and precision in the proposed surrogate model.

\begin{wrapfigure}{r}{4.2cm}
    \centering
    \includegraphics[width=4.2cm]{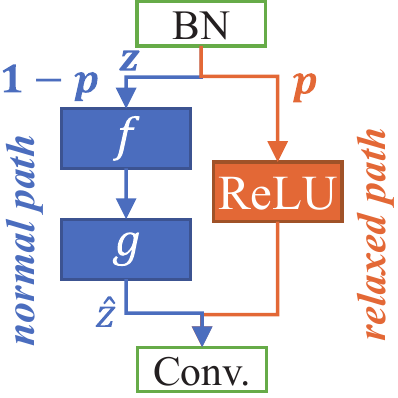}
	\caption{Stochastic relaxation}
	\label{fig:stochastic_relax}
	%\vspace{-2.0em}
\end{wrapfigure}

To investigate this trade-off that is caused by the maximum kernel value $z_{\textrm{M}}$, we trained deep SNNs on various maximum values with an initial $\tau_{0}$.
During the training, we evaluated three kinds of errors: maximum, minimum, and precision errors, which are defined as $z-\hat{z}_{\textrm{M}}$ (if $z>\hat{z}_{\textrm{M}}$), $\hat{z}_{\textrm{m}}-z$ (if $z<\hat{z}_{\textrm{m}}$), and $z-\hat{z}$ (otherwise), respectively.
%As the maximum value was decreased, the maximum error increased, and the minimum and precision errors were decreased (Fig.~\ref{fig:supp_kernel_error} in Supp.).
As the maximum value was decreased, the maximum error increased, and the minimum and precision errors were decreased (for more detailed information, please refer to Appendix).
The total error, which is sum of the three errors, and the representation range are proportional to the initial maximum representation $\hat{z}_{\textrm{M,0}}$, as depicted in Fig.~\ref{fig:tk_error_range}.
Although the cases of larger maximum values $\hat{z}_{\textrm{M,0}}$ have wider expression ranges, they show lower training results, as illustrated in Fig.~\ref{fig:spike_acc_sr_reg}-(a).
This training tendency was based on the fact that the errors by pruned or quantized information, which were caused by the minimum or precision errors, respectively, had a significant effect on the training performance.
Thus, we need to train the surrogate model with narrowed activation distributions to enhance the training performance.
However, the narrower the distribution, such as the case of $\hat{z}_{M,0}$=1, the more activations are pruned or clipped, which leads to significant deterioration in training due to gradient vanishing, especially as SNNs have more layers.

\begin{wrapfigure}{l}{6cm}
    \centering
    \includegraphics[width=6cm]{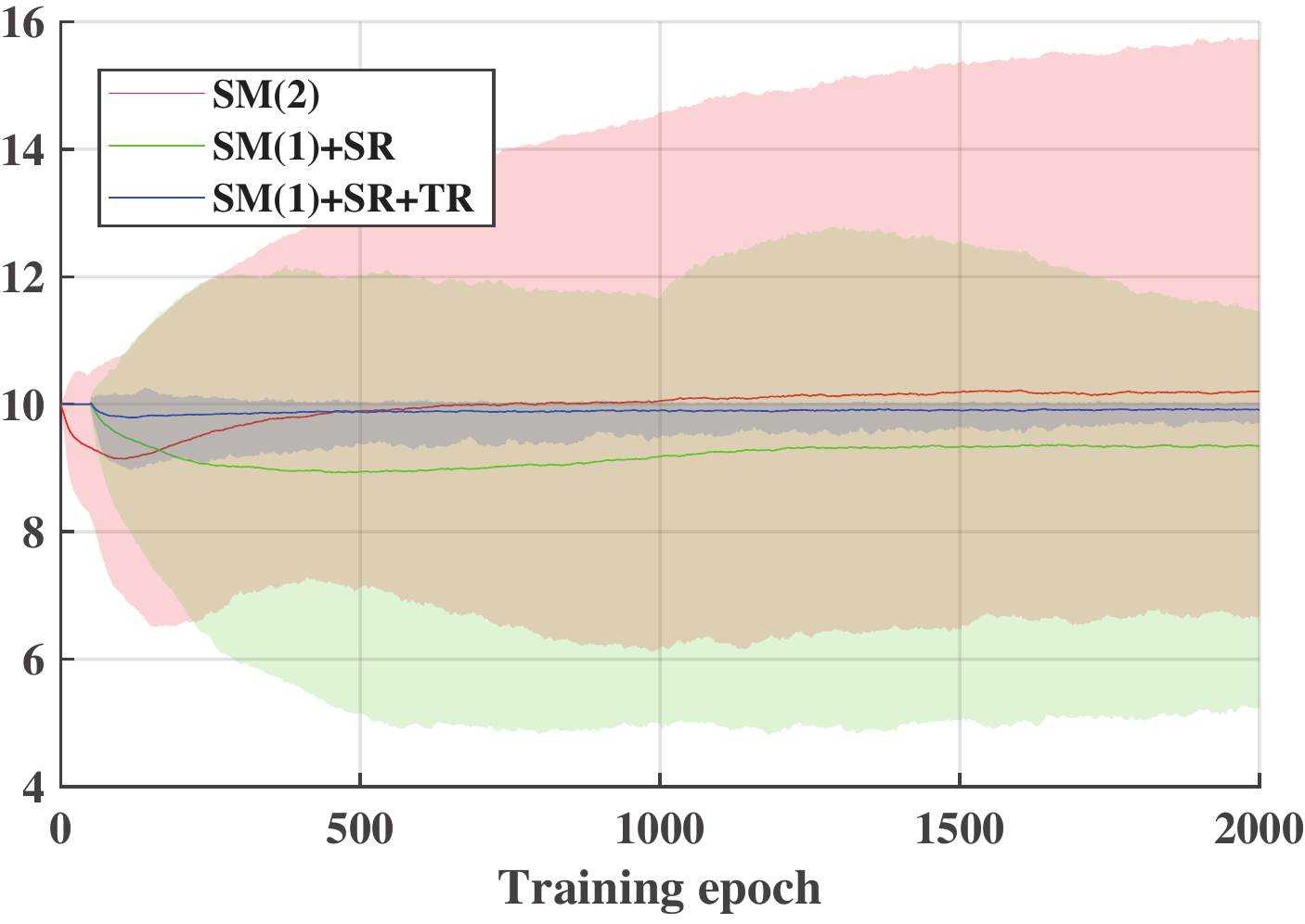}
	\caption{Temporal kernel parameter $\tau$ distribution during training}
	\label{fig:tc_dist}
	\vspace{-0.5em}
\end{wrapfigure}

% stochastic relaxation
To address the aforementioned issue, we propose stochastic relaxation, which lightens the restricted activations on the surrogate model for TTFS coding with a certain probability, as shown in Fig.~\ref{fig:stochastic_relax}.
We used ReLU activation in the relaxed path to overcome the gradient vanishing problem as in conventional DNNs.
In order for the surrogate model to be trained with the characteristics of TTFS coding, we linearly reduced the relaxation effect from the beginning until predefined target training epoch $e_{\textrm{tar}}$.
The probability of stochastic relaxation (SR) is defined as $p(e)=1-(e+1)/e_{\textrm{tar}}$, where $e$ and $e_{\textrm{tar}}$ are training epoch and target epoch of the relaxation, respectively.
In addition, for generalization performance, we applied the relaxation in a layer-wise stochastic manner by using uniformly sampled random variables $r^{l}$ between zero and one, as $p(e) \geq r^{l}$. 
By the proposed relaxation, we were able to train deep SNNs with a narrow representation range, which led to considerable reduction in the errors by the temporal kernel, as depicted in Fig.~\ref{fig:tk_error_range}.
The improved training results of the relaxation are illustrated in Fig.~\ref{fig:spike_acc_sr_reg}-(a).

With the relaxation, we were able to train the SNN, but as the training progressed, the performance was limited due to the divergence of the temporal kernel parameter $\tau$.
The influence of temporal kernel on accuracy becomes small when $\tau$ is diverse (Eq.~\ref{eq:grad_tk}), which results in the insufficient training performance, as depicted in Fig.~\ref{fig:spike_acc_sr_reg}.
To overcome this problem, we propose an initial value-based temporal kernel regularization (TR).
The proposed regularization constrains the kernel parameters based on the initial values with L2 norm as $REG_{\textrm{TR}} = \sum_{l}{(\tau^{l}-\tau_{\textrm{0}}^{l})^2} + \sum_{l}{(t_{\textrm{d}}^{l}-t_{d,\textrm{0}}^{l})^2}$.
We initiated the parameters with empirically founded appropriate values to obtain high training performance with a small number of spikes.
By applying this method, divergence of kernel parameters can be prevented, as shown in Fig.~\ref{fig:tc_dist}, which results in improved training performance with similar number of spikes.
The proposed training methods in this section focus on improving training performance.
Thus, to achieve energy-efficient deep SNNs, a method of reducing the number of spikes is required.

\section{Temporal Kernel-aware Batch Normalization}

To obtain the energy-efficient deep SNNs, we need to reduce the number of spikes, which induce synaptic operations, while maintaining the training performance.
As in previous research~\cite{zheng2020going}, we found the possibility of achieving this goal in the batch normalization that has been widely used to improve generalization performance of training in most conventional DNNs~\cite{ioffe2015batch}.
With the batch normalization, we can provide scaled and shifted normalized activation for each layer.
Based on this fact, the authors in \cite{zheng2020going} proposed threshold-dependent batch normalization, which scales the normalized value by the threshold of spiking neurons to prevent under- and over-activation problem.
However, their empirical methods neither can be applied to TTFS coding with dynamic threshold nor improve the efficiency.

\begin{wrapfigure}{r}{4.3cm}
    \centering
    \includegraphics[width=4.3cm]{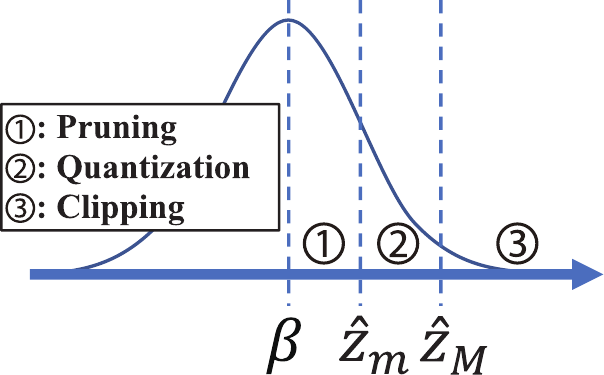}
	\caption{Temporal kernel-aware batch normalization}
	\label{fig:tk_bn}
	\vspace{-1.4em}
\end{wrapfigure}

To reduce the number of spikes by exploiting the batch normalization, we propose temporal kernel-aware batch normalization (TB) in the surrogate model.
The proposed approach utilizes the fact that the decoded activations, which are smaller than the minimum representation of the kernel $\hat{z}_{\textrm{m}}$, are in the pruning region and cannot affect the subsequent layers, as depicted in Fig.~\ref{fig:tk_bn}.
Hence, our intuition of the TB is maximizing the probability of pre-activations $z$ corresponding to the pruning region (\circled{1}), so that the number of spikes is minimized.
To do that, we integrated probability density function (PDF) of the normalized activations by BN from $\beta$ to $\hat{z}_{\textrm{m}}$.
For the computationally efficient integration, we approximated the PDF by Taylor series as
\begin{equation}
\label{eq:pdf_bn}
    p_{\textrm{z}}(\gamma z + \beta) = \exp{(-1/2(\gamma z + \beta)^2)} \approx 1 - 1/2(\gamma x + \beta)^2 \textrm{.}
\end{equation}
With this approximated PDF, we could obtain the integration as
\begin{equation}
\label{eq:integ_pdf_bn}
    P_{\textrm{TB}}(\gamma,\beta)
    =\int_{\beta}^{\hat{z}_{\textrm{m}}} p_{\textrm{B}}(\gamma x + \beta) dz
    \approx 1/2(1/3\gamma^2+\gamma+1)\beta^3
    -1/2\hat{z}_{\textrm{m}}\beta^2
    -1/2(\hat{z}_{\textrm{m}}^2\gamma+2)\beta
    -(1/6\hat{z}_{\textrm{m}}^2\gamma^2-1)\hat{z}_{\textrm{m}} \textrm{.}
\end{equation}
We updated the parameters $\gamma$ and $\beta$ with the gradients of proposed $P_{\textrm{TB}}$ in addition to the conventional training of batch normalization.
The gradients of $P_{\textrm{TB}}$ with respect to $\gamma$ and $\beta$ are explained in Appendix.
Total loss function of the proposed training framework, including TR and TB, is stated as
\begin{equation}
\label{eq:total_loss}
    L = L_{\textrm{CE}} + \lambda_{\textrm{TR}} {REG}_{\textrm{TR}} - \lambda_{\textrm{TB}} {P}_{\textrm{TB}} \textrm{,}
\end{equation}
where $\lambda_{\textrm{TR}}$ and $\lambda_{\textrm{TB}}$ are hyperparameters representing weights of regularization.

%the decoded activation in pruned region 
%increase the probability of normalized activation $z$ in the pruning region
%PDF of normalized distribution
%integration of the PDF between beta and the minimum representation value $\hat{z}_{\textrm{m}}$ of the kernel 
%To get the integral, we approximated the PDF by Taylor series as
%or we simplify the PDF of normalization
%\begin{equation}
%\label{eq:pdf_bn}
    %p_{\textrm{z}}(\gamma z + \beta) = \exp{(-1/2(\gamma z + \beta)^2)} \approx 1 - 1/2(\gamma x + \beta)^2 \textrm{.}
%\end{equation}
%for the computational efficiency, we remove unnecessary coefficients,

%ocal learning based
%ssume BN ~N(0,1), x shift and scaling (gamma, beta)
%to minimize the number of spikes
%> minimize the activations correspoing theporal kernel

%PDF -> integral -> area

%the integral of normal distribution -> 불가능?
%approximate by Taylor series
%-> equation

%reducing the number of spikes
%-> maximize the number of pruned activation 
%-> maximize the area between beta and $\hat{z}_{\textrm{m}}$ on the PDF

%
%pruning by the temporal kernel and BN

%

%$\operatorname{arg\,max}_{\gamma,\beta} P_{\textrm{TB}}(\gamma,\beta)$
%$L_{\textrm{TB}} = -P_{\textrm{TB}}(\gamma,\beta)$

% final loss function
%\begin{equation}
%   L = L_{\textrm{CE}} + \sum_{l}{(\tau^{l}-\tau_{\textrm{0}})^2} + \sum_{l}{(t_{\textrm{d}}^{l}-t_{d,\textrm{0}})^2} + BN loss \textrm{,}
%end{equation}

%\begin{equation}
    %L = L_{\textrm{CE}} + \rho_{\textrm{TR}} {REG}_{\textrm{TR}} + \rho_{\textrm{TB}} {P}_{\textrm{TB}} \textrm{,}
%\end{equation}
%where $\rho_{\textrm{TR}}$ and $\rho_{\textrm{TR}}$ are 

\section{Experimental Results}

\begin{figure}[t]
    \centering
    \includegraphics[width=1.0\linewidth]{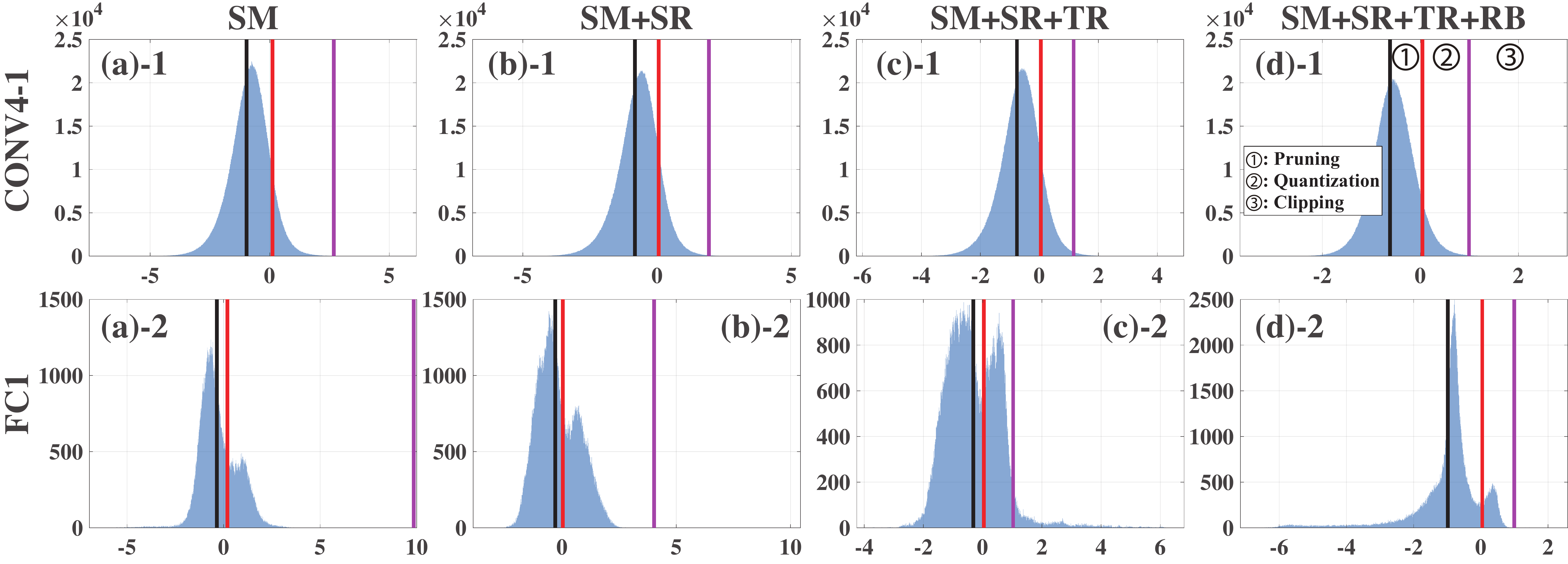}
	%\vspace{-0.6em}
	\caption{Pre-activation $z$ distributions according to the proposed methods. Black, red, and purple vertical lines indicate $\beta$ of batch normalization, the minimum $\hat{z}_{\textrm{m}}$, and maximum $\hat{z}_{\textrm{M}}$ representation of the temporal kernel, respectively (SM: Surrogate Model, SR: Stochastic Relaxation, TR: Temporal kernel Regularization, TB: Temporal kernel-aware Batch normalization).}
	\label{fig:act_dist}
	\vspace{-1.2em}
\end{figure}

We trained deep SNNs, which were based on VGG16, on the CIFAR-10 and CIFAR-100 datasets to evaluate our proposed methods.
We implemented the proposed training framework of deep SNNs on TensorFlow~\cite{tensorflow2015-whitepaper}, which is a well-known deep learning library.
Due to the end-to-end trainable property of the proposed methods, we could exploit the well-defined deep learning framework.
Similarly to other related studies~\cite{zhang2019tdsnn,park2020t2fsnn,han2020deep}, we encoded the input intensity into spike time and used integrated membrane potentials in the output layers as output of SNNs.
We set the initial values of $T$, $t_{\textrm{d,0}}$, $\tau_{0}$ to 32, 0 ,10, respectively, which correspond to $\hat{z}_{\textrm{M,0}}$=1.
Our experiments conducted on NVIDIA TITAN RTX GPU, and each training took more than two and three days on a single GPU for CIFAR-10 and CIFAR-100, respectively.
For more detailed experimental setups, including hyperparameters, please refer to Appendix.

%
%\subsection{Comparison with Other methods for TTFS coding (?)}
%Comparison with Conversion Methods
%\input{6-1_comp_ttfs.tex}
\subsection{Ablation Studies}

%\begin{table}[h]
\begin{figure}[t]
\begin{minipage}{\linewidth}
\begin{minipage}[c]{0.44\linewidth}
    \centering
%    \label{tab:ablation_study}
%    \captionof{table}{Test accuracy and the number of spikes of the proposed methods on CIFAR-10 dataset (time steps=544)}
%    \begin{tabular}{l|cc}
%        \toprule
%        \multirow{2}{*}{Model} & Acc. & \multicolumn{1}{c}{Spikes} \\
%        & (\%) & \multicolumn{1}{c}{($10^4$)} \\
%    	\midrule
%    	SM\footnote{aa}  & 87.43 & 7.23 \\
%    	SM+SR       & 91.05 & 7.31 \\
%    	SM+SR+TR    & 91.83 & 7.31 \\
%    	SM+SR+TR+TB\tmark[b] & \textbf{91.90} & 6.74 \\
%    	SM+SR+TR+TB\tmark[c] & 91.49 & \textbf{4.90} \\
%    	\bottomrule
%    \end{tabular}
    %\begin{tablenotes}\footnotesize
        %\item{$a$}] $\hat{z}_{\textrm{M,0}}$=2 ($\hat{z}_{\textrm{M,0}}$=1 in the other cases) 
        %\item{$b$}] $\lambda_{\textrm{TB}}$=E-5, $c$] $\lambda_{\textrm{TB}}$=E-4 
    %\end{tablenotes}
    
    %
    %\setlength{\textfloatsep}{0pt}
    \ctable[
    pos = H,
    %center,
    caption = {Test accuracy and the number of spikes of the proposed methods on CIFAR-10 dataset (time steps=544)},
    %captionskip = -1.0ex,
    %mincapwdth = \textwidth,
    %width=\columnwidth,
    %width=8cm,
    %pos = H,
    label = {tab:ablation_study},
    %doinside = {\small \def\arraystretch{.8}}
    %doinside = {\footnotesize \def\arraystretch{.8} \setlength{\tabcolsep}{5pt}}
    ]{l|cc}{
        \tnote[a]{$\hat{z}_{\textrm{M,0}}$=2 ($\hat{z}_{\textrm{M,0}}$=1 in the other cases)}
        \tnote[b]{$\lambda_{\textrm{TB}}$=E-5, $^c$ $\lambda_{\textrm{TB}}$=E-4}
        %\vspace{-6.4em}
    }{
        \toprule
        \multirow{2}{*}{Model} & Acc. & \multicolumn{1}{c}{Spikes} \\
        & (\%) & \multicolumn{1}{c}{($10^4$)} \\
    	\midrule
    	SM\tmark[a]  & 87.43 & 7.23 \\
    	SM+SR       & 91.05 & 7.31 \\
    	SM+SR+TR    & 91.83 & 7.31 \\
    	SM+SR+TR+TB\tmark[b] & \textbf{91.90} & 6.74 \\
    	SM+SR+TR+TB\tmark[c] & 91.49 & \textbf{4.90} \\
    	\bottomrule
    }
\end{minipage}
\begin{minipage}[c]{0.53\linewidth}
%\begin{wrapfigure}{r}{5.0cm}
    %\includegraphics[width=1.0\linewidth]{figures/tradeoff_acc_spike.PNG}
    %\includegraphics[width=1.0\linewidth]{figures/spike_acc_tk_bn.PNG}
    \includegraphics[width=1.0\linewidth]{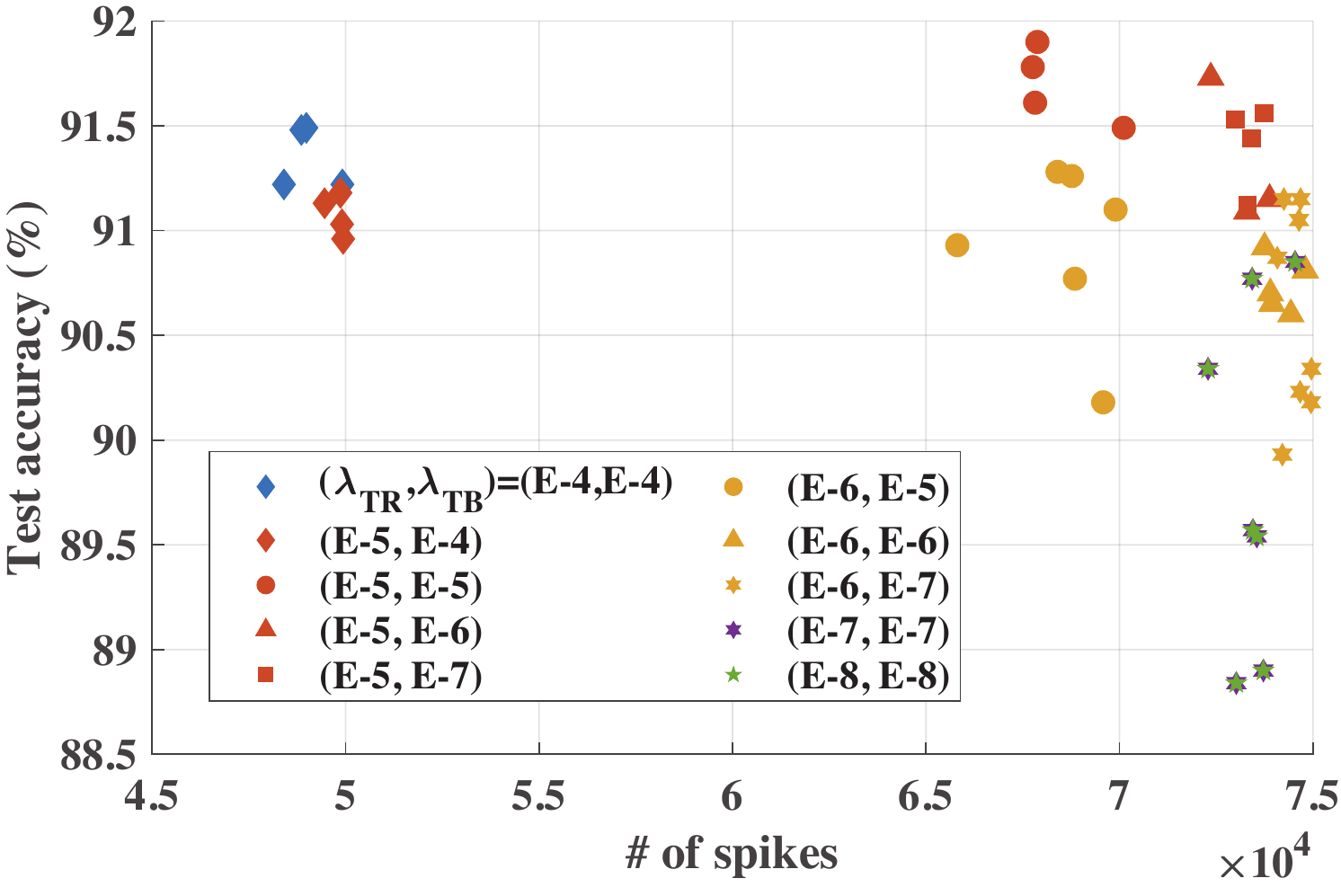}
	\vspace{-2.0em}
	%\captionof{figure}{Test accuracy for the number of spikes in the proposed methods depending on $\lambda_{\textrm{TR}}$ and $\lambda_{\textrm{TB}}$ (VGG-16, CIFAR-10)}
	\captionof{figure}{Test accuracy for the number of spikes in the proposed methods depending on $\lambda_{\textrm{TR}}$ and $\lambda_{\textrm{TB}}$}
	\label{fig:spike_acc_tk_bn}
%\end{wrapfigure}
\end{minipage}
\end{minipage}
%\vspace{-1.0em}
\end{figure}
%\end{table}

%
To evaluate the effect of proposed training methods, we examined the pre-activation values $z$ and the temporal kernel of each layer with a batch of test data after the training.
The results of two layers (CONV4-1, FC1) are depicted in Fig.~\ref{fig:act_dist}, and the results of the remaining layers are in Appendix.
The black, red, and purple vertical lines in Fig.~\ref{fig:act_dist} indicate $\beta$ of batch normalization, the minimum $\hat{z}_{\textrm{m}}$, and maximum representation $\hat{z}_{\textrm{M}}$ of the kernel, respectively.
In the case of surrogate mode with $\hat{z}_{\textrm{M,0}}$=2 (Fig.~\ref{fig:act_dist}-(a)), representation range of the temporal kernel (the length between purple and red vertical bars) was wide, which led to significant errors in quantization and minimum representation, as we discussed in Sec.~\ref{sec:training_methods}.
By applying the stochastic relaxation (Fig.~\ref{fig:act_dist}-(b)), we were able to train the deep SNNs with the temporal kernel.
With the regularization on temporal kernel parameters (Fig.~\ref{fig:act_dist}-(c)), the kernel representation range was smaller, and pre-activations were trained accordingly to the kernel.
Due to the narrowed representation, we improved the accuracy, but the number of spikes increased.
Finally, when the temporal kernel-aware batch normalization was applied (Fig.~\ref{fig:act_dist}-(d)), the deep SNN was trained with a narrower range of kernel and pre-activations.
In addition, the probability of pruning region (\circled{1}) increased while the probability of quantization and clipping regions (\circled{2} and \circled{3}) decreased, which led to significant reduction in the number of spikes.

The training results of the proposed methods in terms of test accuracy and the number of spikes per sample are depicted in Table~\ref{tab:ablation_study} and Fig.~\ref{fig:spike_acc_tk_bn}.
As shown in Table~\ref{tab:ablation_study}, the training performance improved as SR and TR were applied, but the number of spikes increased.
The increased spike count could be reduced by TB. 
By applying TB, with similar accuracy, the number of spikes was reduced by about 7.8\%, and with less than 1\% accuracy degradation, the number of spikes could be reduced by about 33\%, compared to the case of SM+SR+TR.
Fig.~\ref{fig:spike_acc_tk_bn} shows training results according to weights of the proposed methods ($\lambda_{\textrm{TR}}$ and $\lambda_{\textrm{TB}}$).
When $\lambda_{\textrm{TR}}$ is small, the training performance was limited due to enlarged temporal kernel parameters and the gradients that is inversely proportional to $\tau$.
With the proper regularization factor, the training was robustly improved.
In the case of the proposed batch normalization for reducing the number of spikes, the effect was insignificant when the weight $\lambda_{\textrm{TB}}$ was less than a certain value (10\^{}(-5)).
As the weight increased, the number of spikes reduced, and we could achieve significant reductions with slight performance degradation.

\subsection{Comparisons with Other Methods}

\ctable[
pos = t,
center,
caption = {Comparisons with various methods},
%captionskip = -1.0ex,
%mincapwdth = \textwidth,
%width=\columnwidth,
label = {tab:exp_comp_other},
doinside = {\small \def\arraystretch{.9} \addtolength{\tabcolsep}{-1.2pt}}
%doinside = {\footnotesize \def\arraystretch{.8} \setlength{\tabcolsep}{5pt}}
]{l|cc|crr|rr|rr}{
    %@\tnote[a]{Efficiency (Eff.) = \# of Spikes per image / Neurons / Latency}
    %\tnote[a]{spiking density:= \# of spikes per image / (\# of neurons $\cdot$ latency)}
    %\tnote[a]{normalized energy estimation results for each task}
    %\tnote[c]{our experimental results}
    %\tnote[c]{CNN: 12c5-2s-64c5-2s-10}
    \tnote[a]{Energy estimation - normalized to the estimation result of the state-the-of-art research \cite{han2020deep}}
    \tnote[b]{TN: TrueNorth~\cite{merolla2014million}, SN: SpiNNaker~\cite{furber2014spinnaker}, $^c$ DNN-to-SNN conversion}
    \vspace{-3.0em}
}{
    \toprule
    %@\multirow{2}{*}{Methods} & \multicolumn{2}{c}{Neural Coding} & \multirow{2}{*}{Model} & Number of &\multicolumn{2}{c}{Accuracy (\%)} & \multirow{2}{*}{Latency} & \multicolumn{2}{c}{Spikes/image} \\
    %@& Input & Hidden & & Neurons & DNN & SNN & \quad & \# ($10^6$) & Eff. \\
    %Neural & Accuracy & \multirow{2}{*}{Latency} & Spikes & \multicolumn{2}{c}{Normalized Energy}\\
    \multirow{2}{*}{Model} & Training & Neural & Accuracy & \multicolumn{1}{c}{Time} & \multicolumn{1}{c|}{Spike} & \multicolumn{2}{c|}{Spike Evaluation} & \multicolumn{2}{c}{Energy Est.\tmark[a]}\\
    & Method & Coding & (\%) & \multicolumn{1}{c}{Step} & \multicolumn{1}{c|}{($10^6$)} & Rate & Sparsity & \multicolumn{1}{c}{TN\tmark[b]} & \multicolumn{1}{c}{SN} \\
	\midrule

	\multicolumn{10}{l}{CIFAR-10} \\
	\midrule
	%Han et al. 2020~\cite{han2020rmp} & Conv\tnote[c] & Rate & 93.39 & \textbf{512} & 2.612 & 1.82 & 930.91 & 6.01 & 9.01 \\
	Han et al. 2020~\cite{han2020rmp} & Conv$^c$ & Rate & 93.39 & \textbf{512} & 2.612 & 1.82 & 930.91 & 6.01 & 9.01 \\
	%\citet{kim2018deep} & Conv & Phase & 91.21 & 1,500 & 35.196 & 8.36 & 12,543.75 & 74.62 & 117.63 \\
	Kim et al. 2018~\cite{kim2018deep} & Conv & Phase & 91.21 & 1,500 & 35.196 & 8.36 & 12,544 & 75 & 118 \\
	%\citet{park2019fast} & Conv & Burst & 91.41 & 1,125 & 6.920 & 2.19 & 2,466.27 & 15.64 & 23.71 \\
	Park et al. 2019~\cite{park2019fast} & Conv & Burst & 91.41 & 1,125 & 6.920 & 2.19 & 2,466 & 16 & 24 \\
	Han et al. 2020~\cite{han2020deep} & Conv & TSC & \textbf{93.57} & \textbf{512} & 0.193 & 0.13 & 68.86 & 1.00 & 1.00 \\
	Park et al. 2020~\cite{park2020t2fsnn} & Conv & TTFS & 91.43 & 680 & 0.069 & 0.04 & 24.59 & 0.94 & 0.71 \\
	Zhou et al. 2021~\cite{zhou2021temporal} & Direct & TTFS & 92.68 & - & - & - & 62.00 & - & - \\
	\textbf{This work}($\lambda_{\textrm{TB}}$=E-5) & Direct & TTFS & 91.90 & 544 & 0.067 & 0.04 & 23.88 & 0.78 & 0.60 \\
	\textbf{This work}($\lambda_{\textrm{TB}}$=E-4) & Direct & TTFS & 91.49 & 544 & \textbf{0.049} & \textbf{0.03} & \textbf{17.46} & \textbf{0.74} & \textbf{0.54}  \\
	\midrule
	\multicolumn{10}{l}{CIFAR-100} \\
	\midrule
	Han et al. 2020~\cite{han2020rmp} & Conv & Rate & 69.40 & \textbf{512} & 2.048 & 1.43 & 729.77 & 5.42 & 8.07 \\
	%\citet{kim2018deep} & Conv & Phase & 68.66 & 8,950 & 258.408 & 10.29 & 92,095.38 & 618.69 & 979.41 \\
	%\citet{kim2018deep} & Conv & Phase & 68.66 & 8,950 & 258.408 & 10.29 & 92,095 & 618.69 & 979.41 \\
	Kim et al. 2018~\cite{kim2018deep} & Conv & Phase & 68.66 & 8,950 & 258 & 10.29 & 92,095 & 619 & 979 \\
	%\citet{park2019fast} & Conv & Burst & 68.77 & 3,100 & 25.074 & 2.88 & 8,936.30 & 62.65 & 96.60 \\
	%\citet{park2019fast} & Conv & Burst & 68.77 & 3,100 & 25.074 & 2.88 & 8,936 & 62.65 & 96.60 \\
	Park et al. 2019~\cite{park2019fast} & Conv & Burst & 68.77 & 3,100 & 25 & 2.88 & 8,936 & 63 & 97 \\
	Han et al. 2020~\cite{han2020deep} & Conv & TSC & \textbf{70.87} & \textbf{512} & 0.170 & 0.12 & 60.57 & 1.00 & 1.00 \\
	Park et al. 2020~\cite{park2020t2fsnn} & Conv & TTFS & 68.79 & 680 & 0.084 & \textbf{0.04} & 29.94 & 0.99 & 0.79 \\
	\textbf{This work} & Direct & TTFS & 65.98 & 544 & \textbf{0.078} & 0.05 & \textbf{27.80} & \textbf{0.82} & \textbf{0.68} \\
	\bottomrule
}

We assessed our method by comparison with studies on various training methods and neural coding in terms of inference accuracy, latency (time steps), and the number of spikes, as shown in Table.~\ref{tab:exp_comp_other}.
In addition, to evaluate efficiencies, we conducted spike evaluation and energy estimation.
The spike evaluation consists of spike rate~\cite{han2020deep} and sparsity~\cite{zhou2021temporal}, which are defined as $S/(NH)$ and $S/N$, respectively, where $S$, $N$, and $H$ are total number of spikes, neurons, and time steps, respectively.
For the sake of brevity in the notation, we represented these evaluation metrics as percentages in Table~\ref{tab:exp_comp_other}.
To estimate the energy consumption, we adopted the methods in \cite{park2020t2fsnn}, which is defined as $\textrm{E}=S\textrm{E}_{\textrm{d}}+L\textrm{E}_{\textrm{s}}$, where $\textrm{E}_{\textrm{d}}$ and $\textrm{E}_{\textrm{s}}$ are dynamic and static energy coefficients depending on the neuromorphic architecture, respectively.

Rate and TSC coding schemes, which were trained through DNN-to-SNN conversion, showed high accuracy.
However, rate coding generated a much higher number of spikes than TSC due to the utilization of temporal information.
Phase and burst coding methods exploited the temporal information but showed low efficiency.
Studies using TTFS coding generated much smaller number of spikes than other methods.
Among those studies, we achieved the smallest number of spikes with the proposed methods.
Our methods generated only about 25\% spikes compared to TSC, with about 2\% accuracy degradation on CIFAR-10 dataset.
Even considering that the TSC, which utilizes inter-spike-interval (ISI) of two spikes to convey information, our reduction in spikes is significant.
In addition, we could reduce the number of spikes by about 30\% compared to TTFS study, which was trained by the conversion~\cite{park2020t2fsnn}, with a similar accuracy. 

In terms of spike evaluation, the proposed method showed significantly lower spike density compared to other methods.
The spike rate, which means the percentage of neurons that fire a spike on average at every time step, was only 0.03\%, which is about 24\% and 89\% of TSC and TTFS~\cite{park2020t2fsnn}, respectively.
In addition, sparsity, which represents the average spikes per neuron during the inference, is about 17\%, which is about 25\% and 71\% of TSC and TTFS~\cite{park2020t2fsnn}, respectively.
To compare the energy consumption, we normalized each estimated energy consumption based on that of TSC, which is the state-of-the-art method.
The estimation results showed that our proposed methods could reduce the energy consumption up to 54\% and 77\% compared to that of TSC and TTFS~\cite{park2020t2fsnn}, respectively.

%
%- apply previous methods
%----- early firing
%---------  TTFS를 잘 활용하기 위해서는 (layer간의 dependency 줄임, latency 감소) early stage of each time window 에 스파이크가 많이 발생하는 것이 좋다

%\section{Discussion}
%\input{7_discussion.tex}

\section{Discussion and Conclusion}

In this work, we presented a training framework for deep SNNs with TTFS coding to achieve accurate and energy-efficient deep SNNs.
Our proposed methods consist of a computationally efficient surrogate DNN model, robust training algorithms, and temporal kernel-aware batch normalization.
With the proposed approaches, we could exploit advantages in both direct training of SNNs and TTFS coding, which resulted in more accurate inference with smaller number of spikes than in a previous TTFS study~\cite{park2020t2fsnn}.
However, our inference accuracy still is not as high as that of the state-of-the-art method~\cite{han2020deep}.
This is because our study aims to reduce the number of spikes with minimum inference latency in order to maximize deep SNN energy efficiency.
In addition, we set the time window $T$ to 32, which corresponds to 5-bit quantization level.
The state-of-the-art paper did not reveal the information about quantization and utilized two spikes to represent information, which represents more precise information.
Given these facts, we believe that our results cannot be considered insignificant.
Furthermore, the number of spikes has a greater impact on neuromorphic chips.
Considering the restricted resources, deep SNNs should be deployed on multiple neuromorphic cores, which leads to spike movements between the cores being a major bottleneck.
Thus, our study has a greater benefit when considering the constraints of neuromorphic chips.
To overcome the aforementioned limitation in this paper, future work will be expansion of the proposed methods to other applications and improvement on the accuracy of deep SNNs while maintaining the high efficiency.
We expect that our framework for the energy-efficient deep SNNs will enable more complex and larger SNNs.

% ack
%\begin{ack}
%\input{90_ack.tex}
%\end{ack}

% ref
%\section*{References}

%References follow the acknowledgments. Use unnumbered first-level heading for the references. Any choice of citation style is acceptable as long as you are consistent.
%It is permissible to reduce the font size to \verb+small+ (9 point) when listing the references.
%Note that the Reference section does not count towards the page limit.
%\medskip

{
\small
\bibliographystyle{IEEEtran}
\bibliography{91_reference.bib}
}

%%%%%%%%%%%%%%%%%%%%%%%%%%%%%%%%%%%%%%%%%%%%%%%%%%%%%%%%%%%%
\section*{Checklist}

%%%% BEGIN INSTRUCTIONS %%%
%The checklist follows the references.  Please
%read the checklist guidelines carefully for information on how to answer these
%questions.  For each question, change the default \answerTODO{} to \answerYes{},
%\answerNo{}, or \answerNA{}.  You are strongly encouraged to include a {\bf
%justification to your answer}, either by referencing the appropriate section of
%your paper or providing a brief inline description.  For example:
%\begin{itemize}
%  \item Did you include the license to the code and datasets? \answerYes{See Section~\ref{gen_inst}.}
%  \item Did you include the license to the code and datasets? \answerNo{The code and the data are proprietary.}
%  \item Did you include the license to the code and datasets? \answerNA{}
%\end{itemize}
%Please do not modify the questions and only use the provided macros for your
%answers.  Note that the Checklist section does not count towards the page
%limit.  In your paper, please delete this instructions block and only keep the
%Checklist section heading above along with the questions/answers below.
%%%% END INSTRUCTIONS %%%

\begin{enumerate}

\item For all authors...
\begin{enumerate}
  \item Do the main claims made in the abstract and introduction accurately reflect the paper's contributions and scope?
    \answerYes{Please refer to Abstract and Introduction.}
  \item Did you describe the limitations of your work?
    \answerYes{Please refer to Discussion and Conclusion.}
  \item Did you discuss any potential negative societal impacts of your work?
    \answerNA{}
  \item Have you read the ethics review guidelines and ensured that your paper conforms to them?
    \answerYes{Yes, I did.}
\end{enumerate}

\item If you are including theoretical results...
\begin{enumerate}
  \item Did you state the full set of assumptions of all theoretical results?
    \answerNA{}
	\item Did you include complete proofs of all theoretical results?
    \answerNA{}
\end{enumerate}

\item If you ran experiments...
\begin{enumerate}
  \item Did you include the code, data, and instructions needed to reproduce the main experimental results (either in the supplemental material or as a URL)?
    \answerYes{To maintain anonymity, we may not disclose the source code during review process. After the review period, we willingly share the code to reproduce the experimental results.}
  \item Did you specify all the training details (e.g., data splits, hyperparameters, how they were chosen)?
    \answerYes{Please refer to Experimental Results section and Appendix for detailed parameter settings.}
	\item Did you report error bars (e.g., with respect to the random seed after running experiments multiple times)?
    \answerYes{We represented our experiments results with scatter plots as shown in Figs.~\ref{fig:spike_acc_sr_reg} and \ref{fig:spike_acc_tk_bn}.}
    \item Did you include the total amount of compute and the type of resources used (e.g., type of GPUs, internal cluster, or cloud provider)?
    \answerYes{Please refer to Experimental Results section} 
\end{enumerate}

\item If you are using existing assets (e.g., code, data, models) or curating/releasing new assets...
\begin{enumerate}
  \item If your work uses existing assets, did you cite the creators?
    \answerYes{Please refer to Introduction}
  \item Did you mention the license of the assets?
    \answerNA{}
  \item Did you include any new assets either in the supplemental material or as a URL?
    \answerNA{}
  \item Did you discuss whether and how consent was obtained from people whose data you're using/curating?
    \answerNA{}
  \item Did you discuss whether the data you are using/curating contains personally identifiable information or offensive content?
    \answerNA{}
\end{enumerate}

\item If you used crowdsourcing or conducted research with human subjects...
\begin{enumerate}
  \item Did you include the full text of instructions given to participants and screenshots, if applicable?
    \answerNA{}
  \item Did you describe any potential participant risks, with links to Institutional Review Board (IRB) approvals, if applicable?
    \answerNA{}
  \item Did you include the estimated hourly wage paid to participants and the total amount spent on participant compensation?
    \answerNA{}
\end{enumerate}

\end{enumerate}
%%%%%%%%%%%%%%%%%%%%%%%%%%%%%%%%%%%%%%%%%%%%%%%%%%%%%%%%%%%%

\newpage
\appendix

\section*{Appendix}
\renewcommand\thesection{\Alph{section}}
\setcounter{section}{0}

\section{Implementation TTFS coding in deep SNNs}

\begin{figure}[h]
    \centering
    \includegraphics[width=0.9\linewidth]{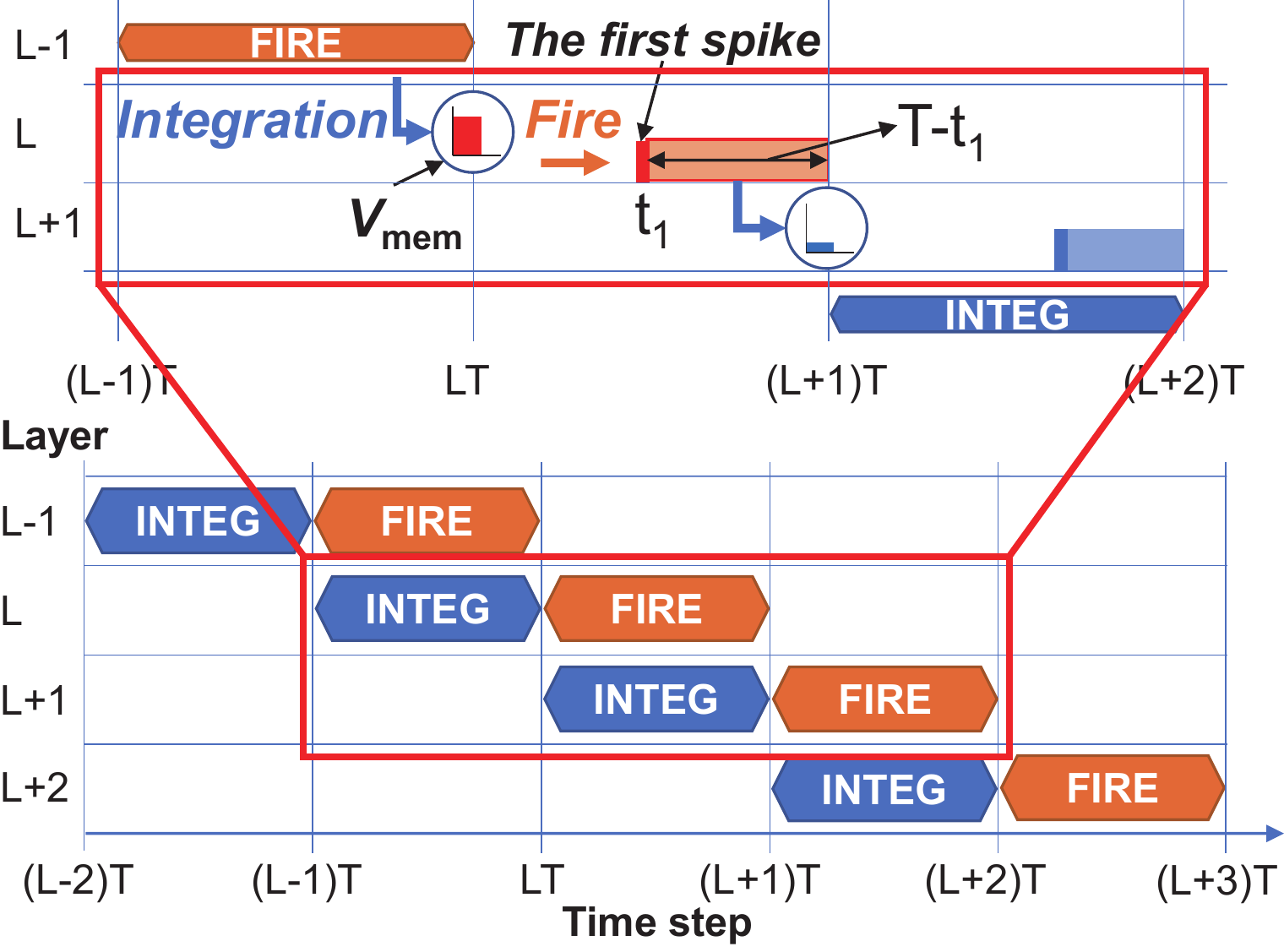}
	%\caption{TTFS coding implementation in deep SNNs (\textcolor{red}{figure should be modified, rate coding, ISI?, TTFS coding 설명, 본문 reference t 기호 그림에 나타나게, 뉴럴 코딩 그림에 나타나게})}
	\caption{TTFS coding implementation in deep SNNs. (INTEG: INTEGration phase, FIRE: FIRE phase)}
	\label{fig:app_TTFS}
\end{figure}

As we mentioned in Section~\ref{sec:snn}, TTFS coding uses the time of the first spike to represent information, and thus can convey information with very few spikes.
For this reason, TTFS coding has been regarded a promising neural coding for energy-efficient SNNs.
As shown in Fig.~\ref{fig:app_TTFS}, it can be implemented in two phases in deep SNNs: integration (INTEG) and fire (FIRE) phases~\cite{zhang2019tdsnn,park2020t2fsnn}.
To represent and transmit information with only one spike, each layer has the two phases, and there are dependencies between each layer's phases.
In the integration phase, the spike of the previous layer is transmitted through the synaptic weight, and the value transmitted to the membrane potential is accumulated into the internal state (membrane potential, $V_{\textrm{mem}}$ in Fig.~\ref{fig:app_TTFS}).
This phase contains the decoding process of TTFS coding.
The fire phase, which corresponds to the encoding process, generates spikes based on the accumulated internal state in the integration phase.

In order to efficiently utilize TTFS coding, it is advantageous for earlier spikes to convey more critical information.
As shown in Fig.~\ref{fig:app_TTFS}, when the first spike time is $t_1$, the corresponding information of the spike is the difference between the pre-defined time window $T$ and the spike time ($T$-$t_1$).
Since TTFS coding expresses information with the time of the first spike, only one spike is used for each neuron, and spikes after the first spike cannot affect the subsequent layers.
Thus, to minimize the number of spikes, a neuron, which generates a spike, enters a sufficiently long refractory period so that no more spike occur from the neuron~\cite{zhang2019tdsnn,park2020t2fsnn}.
If the refractory period is larger than the given time window $T$ in each layer, it is not difficult to prevent the subsequent spike from occurring.
The efficient implementation of TTFS coding in deep SNNs can significantly reduce the number of spikes compared to other neural coding methods, but this implementation comes at the cost of long inference times due to the dependency between adjacent layers.

%2.1 SNN
%how to implement TTFS coding 
%
%reset function - sufficiently long refractory period after the first spike generates
%
%수식 추가해서 설명
%refractory period > time window of each layer
%refractory period 동안 empirically negative values
%max of the synaptic kernel $\kappa$ is sufficient
%

%
%우리가 본문에서 언급했듯이, TTFS 코딩은 첫 번째 스파이크 시각을 사용하여 정보를 표현하하고, 이에따라 매우 적은 스파이크로 정보를 전달할 수 있다.
%TTFS coding은 deep SNN에 그림과 같이 구현될 수 있다.
%이 방법은 각 레이어가 integration 과 fire 두 phase를 갖는 방법이다.
%integration phase에서는 이전 레이어의 스파이크가 synaptic weight를 통해 전달되어 membrane potential에 전달된 값이 누적되는 과정으로, TTFS 코딩의 decoding 과정이 포함되어 있다.
%fire phase는 integration phase에서 축적된 값을 기반으로 spike를 생성하는 과정으로, encoding 과정을 포함하고 있다.
%그림에서 볼 수 있듯이, 첫 spike time 이 t1인 경우, time window와 그 스파이크의 차이 T-t1에 해당하는 정보를 표현하게 된다.
%TTFS coding은 처음 스파이크 시각으로 정보를 표현하기 때문에, 각 뉴론당 최대 스파이크 하나만 사용한다.
%첫 스파이크 이후의 스파이크 들은 정보 전달에 활용되지 않기 때문에, refractory 등을 활용하여 스파이크를 한번 생성한 뉴런은 더이상 스파이크를 생성하지 못하게 하여 TTFS coding에서 발생하는 스파이크 수를 최소화 할 수 있다.
%스파이크가 발생하지 않는 refractory를 주어진 time window보다 크게하면 어렵지 않게 뒤 이은 스파이크 발생을 막을 수 있다.

\newpage
\section{Temporal kernel}

\begin{figure}[h]
    \centering
    \includegraphics[width=1.0\linewidth]{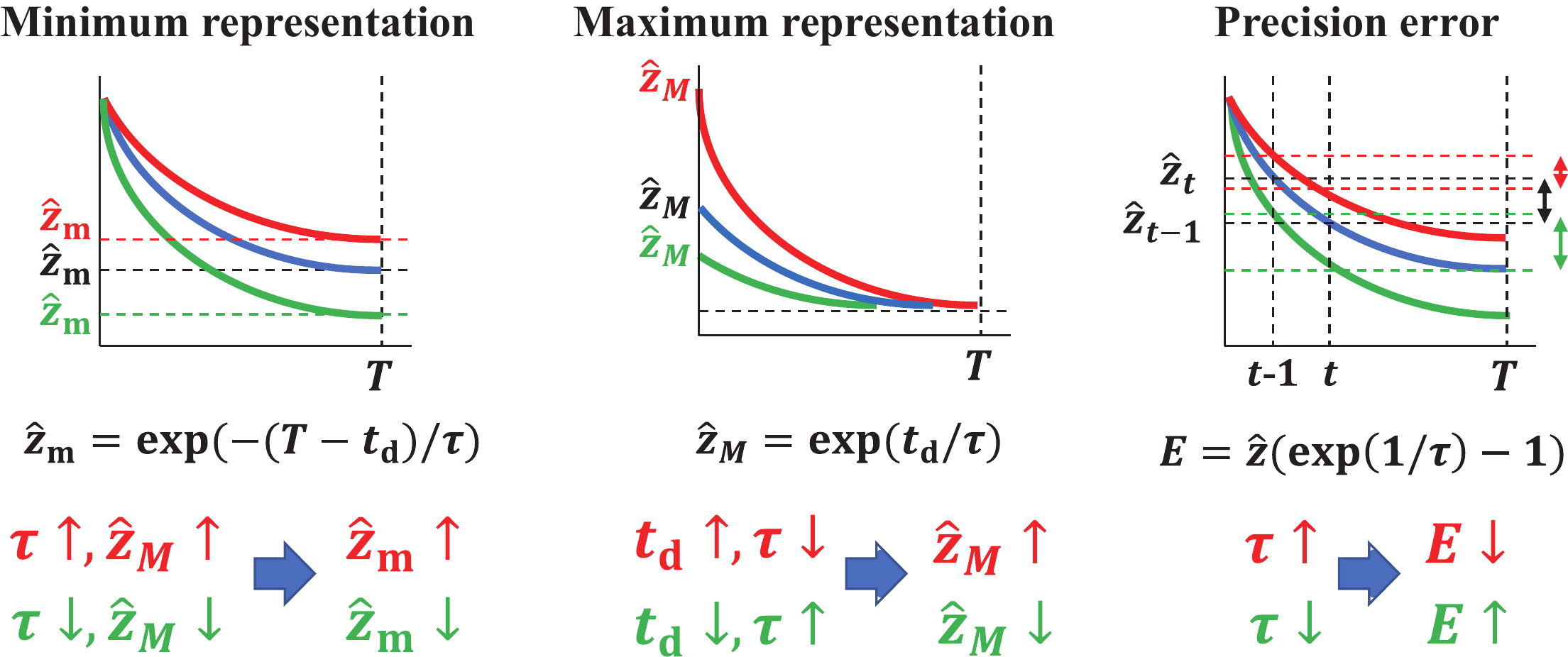}
	\caption{The effect of temporal kernel parameters on the three types of temporal kernel errors}
	\label{fig:app_temporal_kernel}
\end{figure}

The temporal kernel of Eq.~\ref{eq:kernel} is an important factor in the efficient implementation of TTFS coding in deep SNNs.
We adopted an exponentially decaying kernel function with two trainable parameters: $\tau$ and $t_{\textrm{d}}$, as in \cite{park2020t2fsnn}.
As we defined in Section.~\ref{sec:training_methods}, there are three types of errors by the kernel: minimum and maximum representation; and precision error, as depicted in Fig.~\ref{fig:app_temporal_kernel}.
These errors are defined according to the pre-activation $z$ as
\begin{align}
    Error(z) 
    & = z-\hat{z}_{\textrm{M}} & & (\textrm{if } z>\hat{z}_{\textrm{M}} \textrm{, maximum representation error})\\
    & = \hat{z}_{\textrm{m}}-z & & (\textrm{if } z<\hat{z}_{\textrm{m}} \textrm{, minimum representation error}) \\
    & = z-\hat{z} & & (\textrm{otherwise, precision error}) \textrm{.}
\end{align}
The maximum and minimum representation errors are caused by clipping and pruning of the temporal kernel, respectively.
The precision error is caused by the quantization that occurs when representing the spike time.
As shown in Fig.~\ref{fig:app_temporal_kernel}, these errors are greatly affected by the kernel parameters ($\tau$, $t_{\textrm{d}}$).
In order for the kernel to cover a wide range of activation, it is beneficial to increase the representation length, which is defined the length between the maximum and minimum representation.
To achieve this with a certain time window $T$, $\tau$ should decrease, but the reduction in the parameter increases the precision error, which leads to degradation in training performance.
%In addition, there is a trade-off between maximum and minimum representation.
In addition, if there are lots of clipped and pruned activation by the kernel, the training of deep SNNs becomes difficult due to the gradient vanishing.
Thus, it is desirable for deep SNNs to be trained with a narrow range of activation.

\begin{figure}[t]
    \centering
    \includegraphics[width=1.0\linewidth]{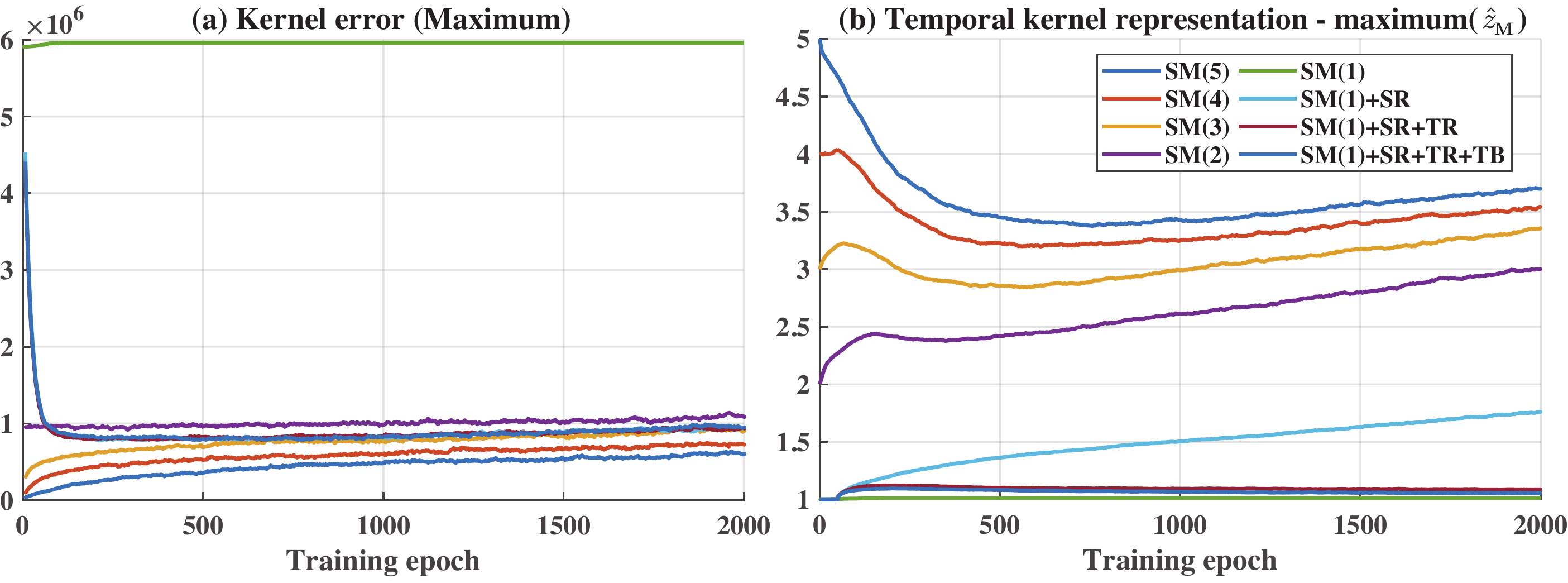}
	\caption{(a) Temporal kernel error by the maximum representation and (b) the maximum representation value during training depending on various configurations. The values in the parenthesis indicate the initial maximum kernel value $\hat{z}_{\textrm{M},0}$ (SM: Surrogate Model, SR: Stochastic Relaxation, TR: Temporal kernel parameters Regularization, TB: Temporal kernel-aware Batch normalization)}
	\label{fig:app_tk_max}
\end{figure}

\begin{figure}[h]
    \centering
    \includegraphics[width=1.0\linewidth]{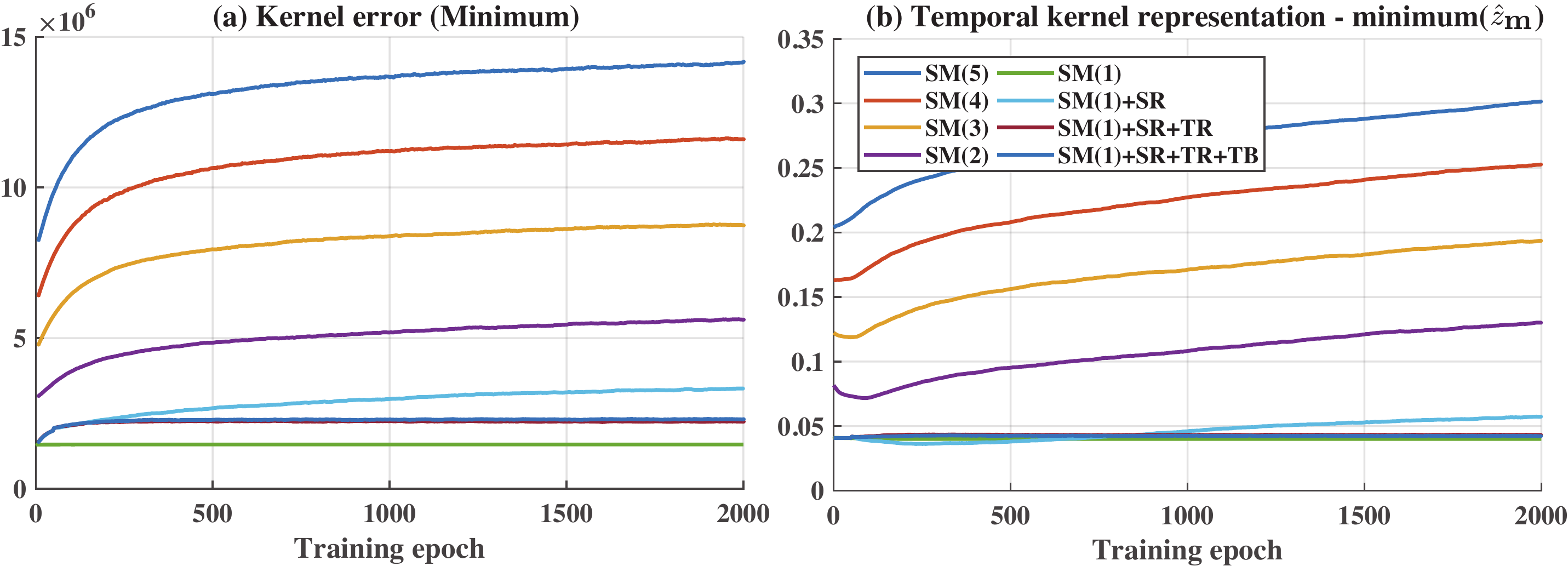}
	\caption{(a) Temporal kernel error by the minimum representation and (b) the minimum representation value during training depending on various configurations. The values in the parenthesis indicate the initial maximum kernel value $\hat{z}_{\textrm{M},0}$ (SM: Surrogate Model, SR: Stochastic Relaxation, TR: Temporal kernel parameters Regularization, TB: Temporal kernel-aware Batch normalization)}
	\label{fig:app_tk_min}
\end{figure}

\begin{figure}[H]
    \centering
    \includegraphics[width=1.0\linewidth]{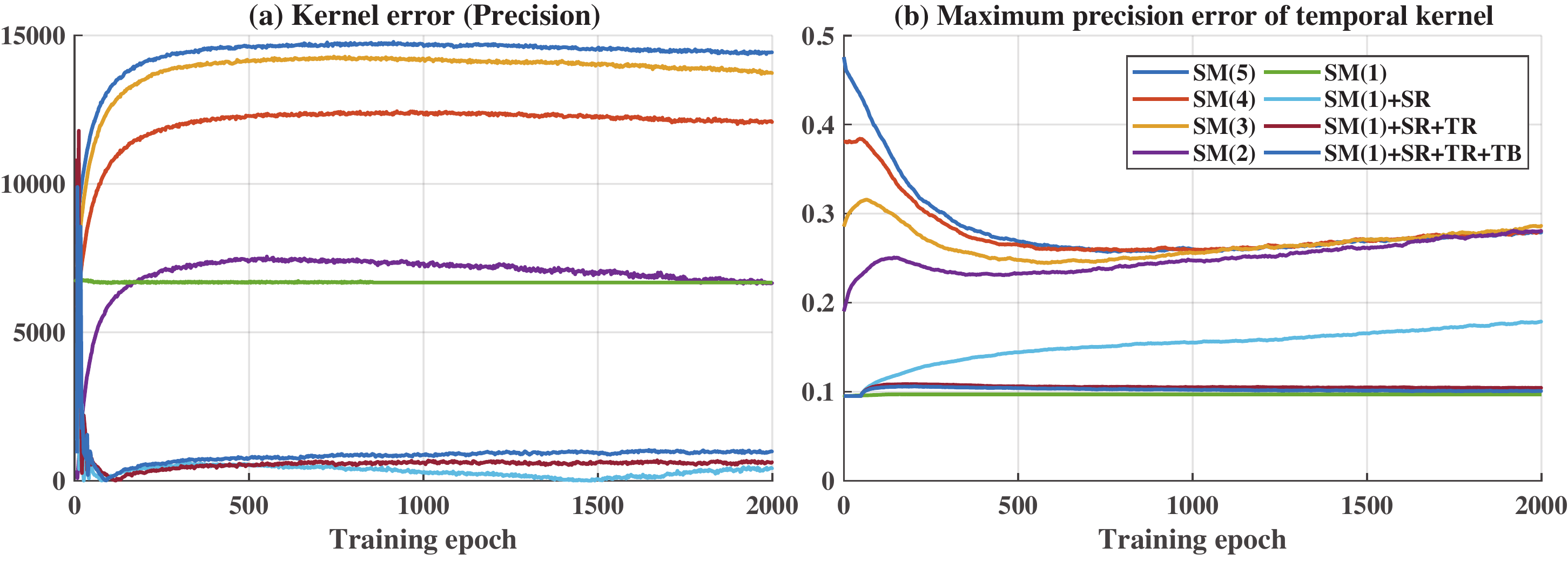}
	\caption{(a) Temporal kernel error by the precision error and (b) the maximum precision error during training depending on various configurations. The values in the parenthesis indicate the initial maximum kernel value $\hat{z}_{\textrm{M},0}$ (SM: Surrogate Model, SR: Stochastic Relaxation, TR: Temporal kernel parameters Regularization, TB: Temporal kernel-aware Batch normalization)}
	\label{fig:app_tk_pre}
\end{figure}

The maximum representation and the temporal kernel error caused by the value are depicted in Fig.~\ref{fig:app_tk_max}.
As the initial maximum representation $\hat{z}_{\textrm{M},0}$ became smaller, the maximum error increased.
In particular, when the initial value was 1, this error increased rapidly.
When the proposed methods (stochastic relaxation (SR) and regularization (TR)) were applied, a reduced maximum kernel error was achieved even with a small maximum expression.
The kernel error by the minimum expression and the minimum expression value are shown in Fig.~\ref{fig:app_tk_min}.
The minimum expression of the kernel was inversely proportional to the initial maximum representation.
When the proposed method is applied, the minimum expression value of the kernel became smaller.
The precision error and the maximum precision error depending on the kernel are shown in Fig.~\ref{fig:app_tk_pre}.
The maximum precision error is defined as
\begin{equation}
\label{eq:app_max_prec_error}
    \exp(t_{\textrm{d}}/\tau)-\exp((t_{\textrm{d}}-1)/\tau) \textrm{.}
\end{equation}
The precision error of the kernel tends to be inversely proportional to the initial maximum representation.
With the proposed methods, we could reduce the precision error significantly.
Through these results, we can improved the training performance with a narrow range of activations by using the proposed methods.

%5. temporal kernel tau distribution during training

%3. surrogate model
%- M(1) 포함된 error
%- training / val loss
%- tk parameters training dist

%\textcolor{red}{activation distribution according to initial clip range -> supp?}

%4. training methods
%maximum precision error in the temporal kernel
%between kernel values t=0 and t=1, because the kernel exponentially decays
%error max = $\exp(t_{\textrm{d}}/\tau)-\exp((t_{\textrm{d}}-1)/\tau)$

%%$\hat{z}_{\textrm{M}}$ = $\exp(t_{\textrm{d}}/\tau)$
%%$\hat{z}_{\textrm{m}}$ = $\exp(-(T-t_{\textrm{d}})/\tau)$

%% trade off -> training performance / efficiency (spikes)
%%maximum - caused by over activation
%%minimum - caused by under activation
%%and precision errors
%%maximum and minimum representation, respectively.

%%temporal kernel's representation range
%%$[\exp(-T/\tau),1]z_{\textrm{max}}^{l}$
%%$[\exp(-T/\tau),1]z_{\textrm{M}}^{l}$
%%$z_{\textrm{m}}^{l}$

%

%\textbf{< Training tendency of the temporal kernel >}

%increment in early stage

%- large tc -> constant, leads to reducing uncertainty

%- help reducing bias in training?

%\textbf{< proposed method for spike loss, regularization >}
%prove the proposed proxy(estimation) of the number of spikes

%prev 버전
%tau 넓은 분포 학습, 발산
%레이어별 큰 분포가 generalization issue에 영향 미치는지 정량적 분석 있으면 좋을 듯
%발산하는 특정 layer가 information transmission (propagation) bottleneck 되는 건지?

%좁은 activation (spike time) 분포로 학습 
%장점:
%- quantization error를 줄일 수 있음 (precision error 감소)
%- important (critical) information을 early time 에 전달 할 수 있음
%---> early firing 등을 활용하여 layer의 dependnecy로 인해 발생하는 increased latency 문제를 완화할 수 있음

\newpage
\section{Temporal kernel-aware batch normalization}

\begin{figure}[h]
    \centering
    \includegraphics[width=1.0\linewidth]{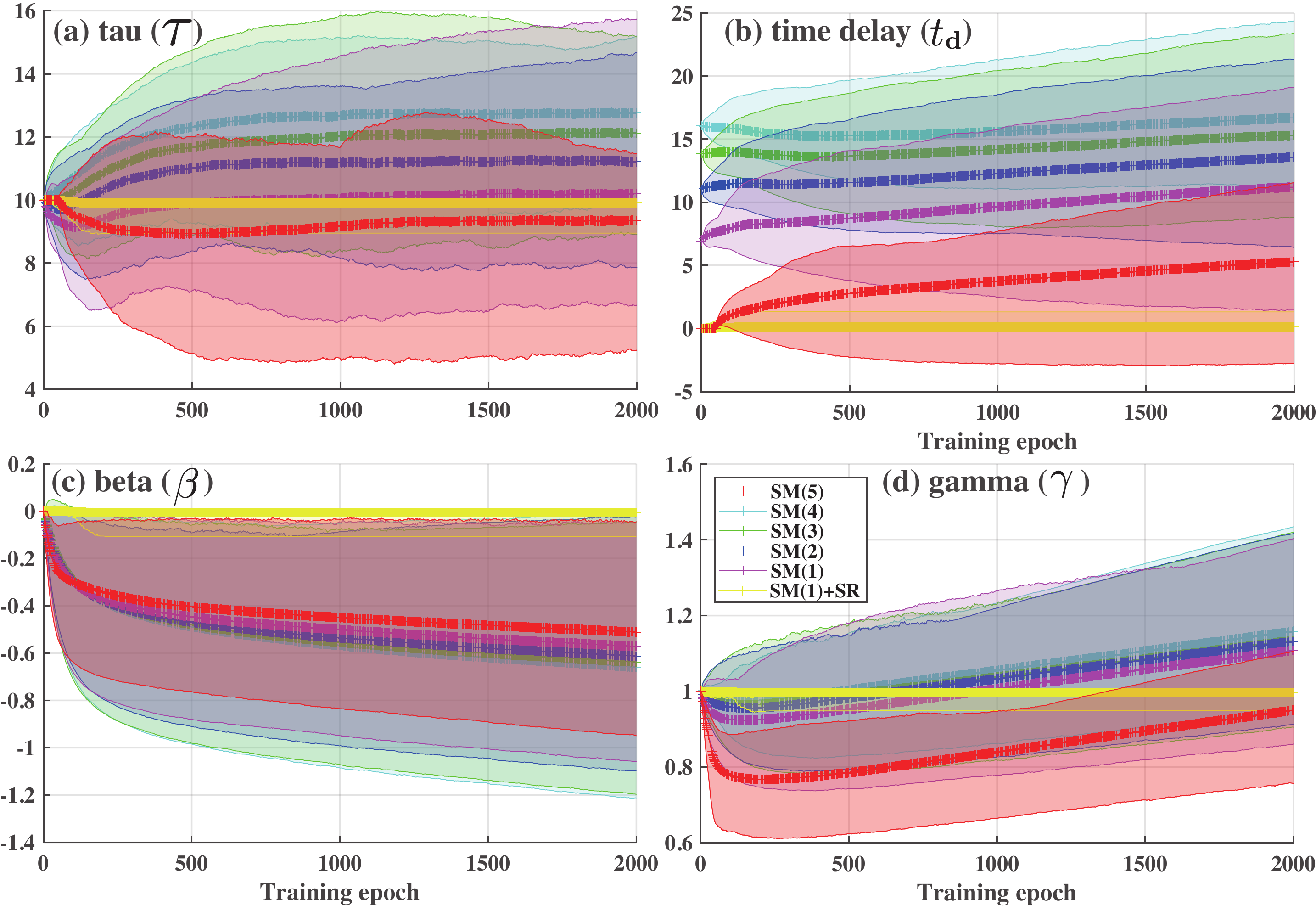}
	\caption{Distributions of parameters in temporal kernel ($\tau$, $t_{\textrm{d}}$)and batch normalization ($\beta$, $\gamma$) depending on the initial maximum kernel value $\hat{z}_{\textrm{M},0}$ during training deep SNNs (VGG-16, CIFAR-10) (SM: Surrogate Model, SR: Stochastic Relaxation)}
	\label{fig:app_training_para_sm}
\end{figure}

\begin{figure}[H]
    \centering
    \includegraphics[width=1.0\linewidth]{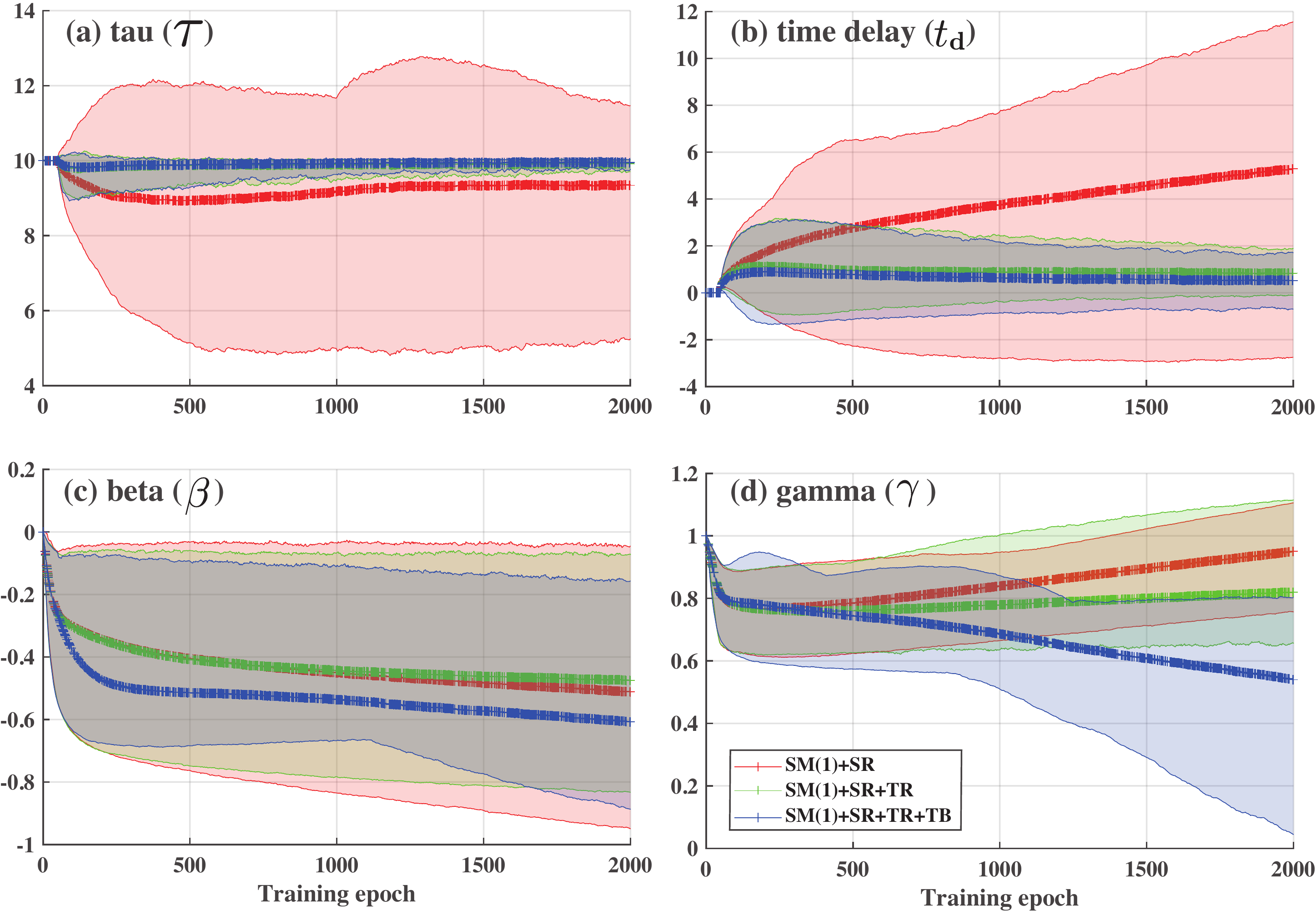}
	\caption{Distributions of parameters in temporal kernel ($\tau$, $t_{\textrm{d}}$)and batch normalization ($\beta$, $\gamma$) depending on various configurations of the proposed methods during training deep SNNs (VGG-16, CIFAR-10) (SM: Surrogate Model, SR: Stochastic Relaxation, TR: Temporal kernel parameters Regularization, TB: Temporal kernel-aware Batch normalization)}
	\label{fig:app_training_para_prop}
\end{figure}

In this work, to improve the energy efficiency of deep SNNs, we proposed temporal kernel-aware batch normalization.
This method reduced the number of spikes in TTFS coding, and thus enabled energy-efficient SNNs.
We utilized the PDF of pre-activation $z$, which is the output of batch normalization, to obtain reduced number of spikes.
For the feasibility and efficiency of the calculation, we approximated the PDF as second-order polynomial by Taylor series.
Through this approximation, we could get the proxy function of the number of spikes in deep SNNs as
\begin{align}
\label{eq:app_integ_pdf_bn}
    P_{\textrm{TB}}(\gamma,\beta)
    & =\int_{\beta}^{\hat{z}_{\textrm{m}}} p_{\textrm{B}}(\gamma x + \beta) dz \nonumber \\
    & \approx 1/2(1/3\gamma^2+\gamma+1)\beta^3
    -1/2\hat{z}_{\textrm{m}}\beta^2
    -1/2(\hat{z}_{\textrm{m}}^2\gamma+2)\beta
    -(1/6\hat{z}_{\textrm{m}}^2\gamma^2-1)\hat{z}_{\textrm{m}} \textrm{.}
\end{align}
The gradients of the proposed batch normalization with respect to $\beta$ and $\gamma$ is derived as 
\begin{equation}
\label{eq:app_grad_beta}
    \frac{\partial{P_{\textrm{TB}}}}{\partial{\beta}}
    = 3/2(1/3\gamma^2+\gamma+1)-\hat{z}_{\textrm{m}}\beta-1/2(\gamma\hat{z}_{\textrm{m}}^2+2) \textrm{ and }
\end{equation}
\begin{equation}
\label{eq:app_grad_gamma}
    \frac{\partial{P_{\textrm{TB}}}}{\partial{\gamma}}
    =-1/3(\hat{z}_{\textrm{m}}^3-\beta^3)\gamma-1/2(\hat{z}_{\textrm{m}}^2-\beta^2)\beta \textrm{, }
\end{equation}
respectively.
The distributions of parameters in the temporal kernel ($\tau$, $t_{\textrm{d}}$) and batch normalization ($\beta$, $\gamma$) depending on the proposed methods are depicted in Figs.~\ref{fig:app_training_para_sm} and \ref{fig:app_training_para_prop}.

\newpage
\section{Experiments}

\ctable[
pos = h,
center,
caption = {Hyperparameters},
%captionskip = -1.0ex,
%mincapwdth = \textwidth,
%width=\columnwidth,
%width=8cm,
%pos = H,
label = {tab:app_hyperpara},
%doinside = {\small \def\arraystretch{.8}}
%doinside = {\footnotesize \def\arraystretch{.8} \setlength{\tabcolsep}{5pt}}
]{lcc}{
    %\tnote[a]{$\hat{z}_{\textrm{M,0}}$=2 ($\hat{z}_{\textrm{M,0}}$=1 in the other cases)}
    %\tnote[b]{$\lambda_{\textrm{TB}}$=E-5, $^c$ $\lambda_{\textrm{TB}}$=E-4}
    \tnote[a]{SR: Stochastic Relaxation}
    \tnote[b]{TR: Temporal kernel parameters Regularization}
    \tnote[c]{TB: Temporal kernel-aware Batch normalization}
    %\vspace{-6.4em}
}{
    \toprule
    Description & Value & Symbol \\
	\midrule
	Model & VGG-16 & - \\
	Datasets & CIFAR-10 / CIFAR-100 & - \\
    Optimizer & ADAM & - \\
    Learning rate & 0.001 & - \\
    Weight regularization & L2 norm. & - \\
    Training batch size & 512 & - \\
    Training epochs & 2,000 / 3-5,000 & - \\
    target training epoch for SR\tmark[a] & 50 / 500-1,000 & $e_{\textrm{tar}}$ \\
    Weight for TR\tmark[b] & E-5 & $\lambda_{\textrm{TR}}$ \\
    Weight for TB\tmark[c] & E-5 & $\lambda_{\textrm{TB}}$ \\
    Time window & 32 & $T$ \\
    Initial threshold & 1.0 & $\theta_0$ \\
	\bottomrule
}

\begin{figure}[h]
    \centering
    \includegraphics[width=1.0\linewidth]{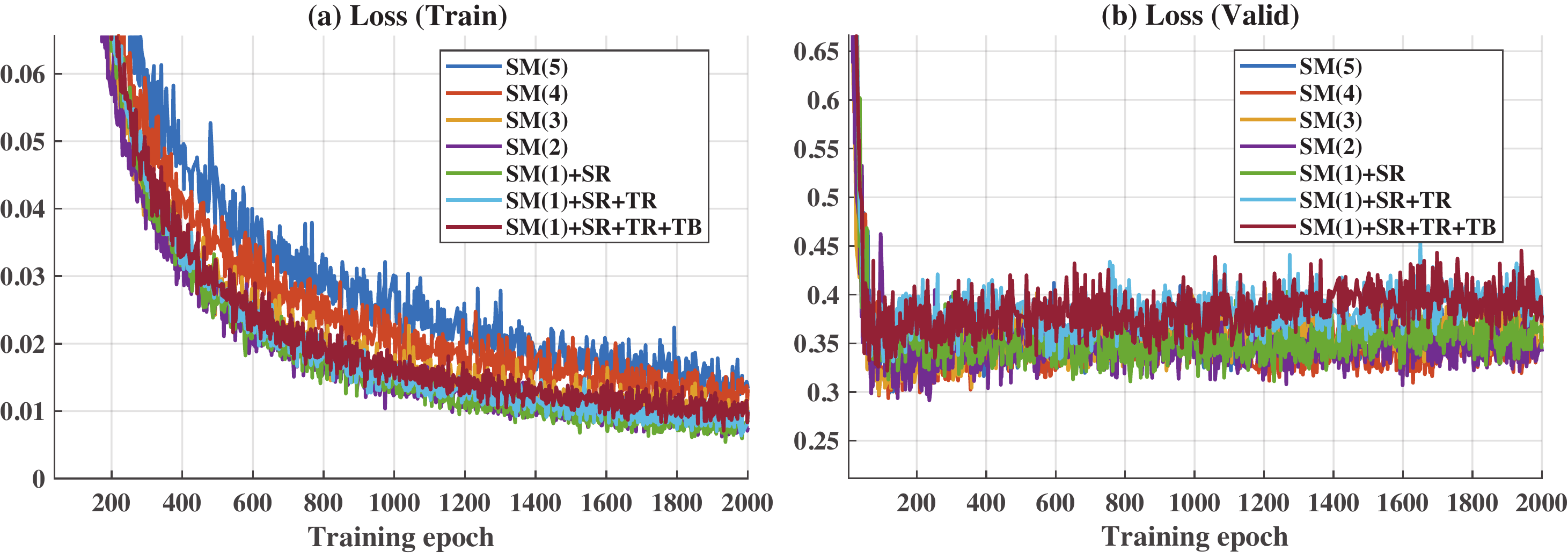}
	\caption{Training and validation loss depending on various configurations}
	\label{fig:app_loss}
\end{figure}

\begin{figure}[h]
    \centering
    \includegraphics[width=1.0\linewidth]{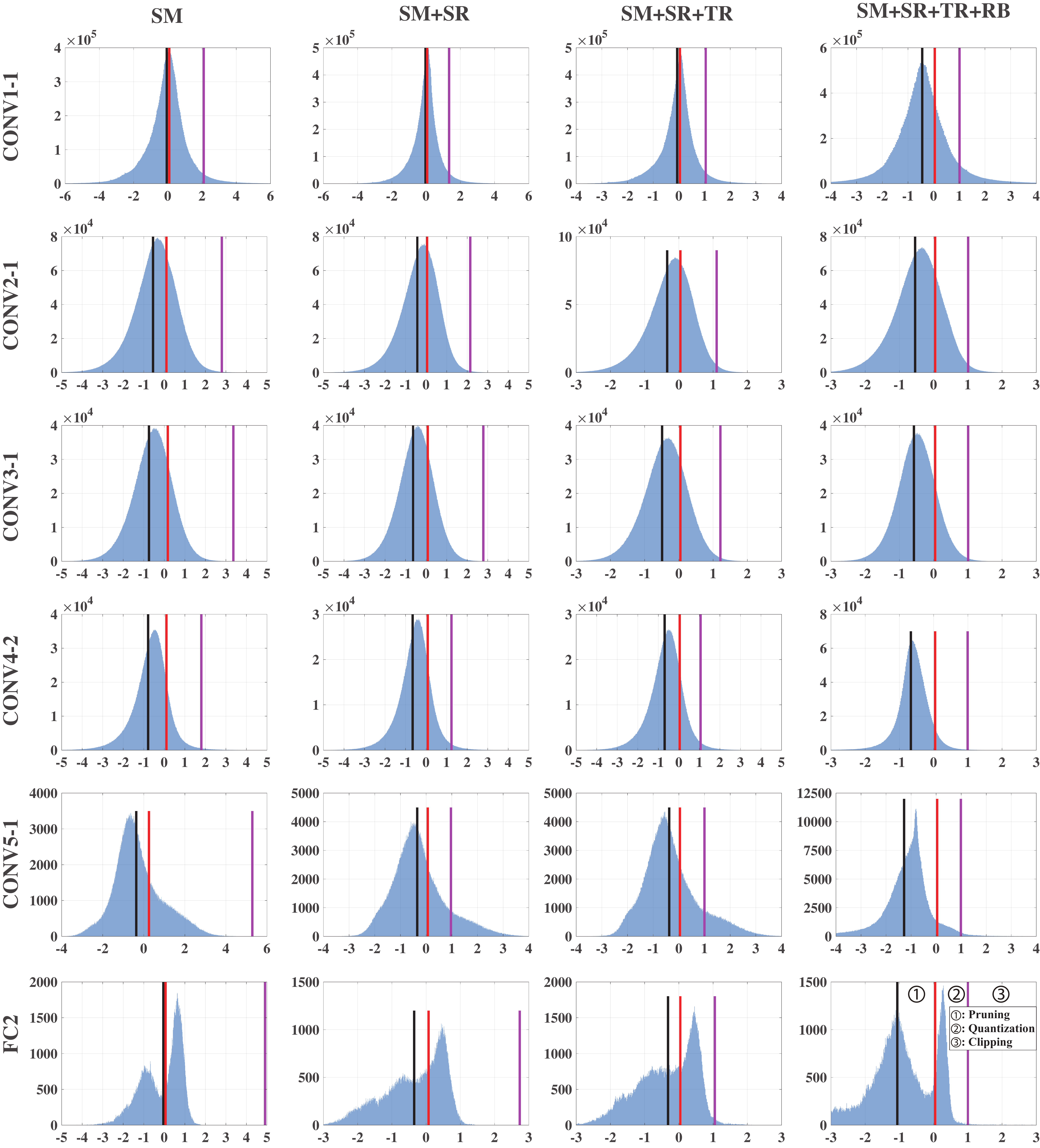}
	\caption{Pre-activation $z$ distributions according to the proposed methods. Black, red, and purple vertical lines indicate $\beta$ of batch normalization, the minimum $\hat{z}_{\textrm{m}}$, and maximum $\hat{z}_{\textrm{M}}$ representation of the temporal kernel, respectively (SM: Surrogate Model, SR: Stochastic Relaxation, TR: Temporal kernel Regularization, TB: Temporal kernel-aware Batch normalization).}
	\label{fig:app_act_dist}
\end{figure}

Hyperparameters used in this study are listed in Table~\ref{tab:app_hyperpara}.
The training and validation loss depending on various configurations are depicted in Fig.~\ref{fig:app_loss}.
We can validate the training performance of the proposed approaches with the training and validation loss.
The distributions of pre-activation $z$ according to the proposed methods are shown in Fig.~\ref{fig:app_act_dist}.
As in Fig.~\ref{fig:act_dist}, we can confirm the effect of the proposed batch normalization on the pre-activation distributions.
We also observe that the pre-activations in the pruned region (\circled{1}) increases by applying the temporal kernel-aware batch normalization, which leads to reduce the number of spikes in deep SNNs.
The effect of the proposed batch normalization is greater as it gets closer to the output layer.

%5. comparison other studies
%- T2FSNN의 EF 적용 성능 비교, 각 layer 별 the first spike time 비교

%

\end{document}